\documentclass[preprint,12pt]{elsarticle}

\usepackage{graphicx}
\graphicspath{{figures/}} 

\usepackage{algorithm}
\usepackage{algpseudocode}
\usepackage[left=2cm, right=2cm, top=2cm, bottom=2cm]{geometry}

\usepackage{booktabs}
\usepackage{tabularray}
\usepackage{array}
\usepackage{multirow}
\usepackage{caption}
\usepackage{tabularx}

\usepackage{amssymb}
\usepackage{amsmath}

\usepackage{url}

\usepackage{ragged2e}
\usepackage{siunitx}
\usepackage{xcolor}
\usepackage{marvosym}
\usepackage{pifont}

\usepackage[utf8]{inputenc}
\usepackage[T1]{fontenc}

\usepackage[colorlinks=true,linkcolor=blue,citecolor=blue,urlcolor=blue]{hyperref}
\usepackage{cleveref}

\usepackage{float}

\usepackage{lineno}
\usepackage{etoolbox}
\AtBeginEnvironment{document}{}
\newcommand{\startmainlinenumbers}{\begin{linenumbers}}
\newcommand{\stopmainlinenumbers}{\end{linenumbers}}

\newcommand{\mysep}{;}

\journal{Automation in Construction}

\begin{document}

\begin{frontmatter}

\title{From Geometric Labels to Semantic Understanding of Indoor Building Components Using Multimodal Large Language Models}

\author[first]{Shuju Jing}
\author[second]{Chao Yin\corref{cor1}}
\cortext[cor1]{Corresponding author.}
\ead{cyinac@connect.ust.hk}

\affiliation[first]{organization={School of Qilu Transportation, Shandong University},
            city={Jinan},
            postcode={250002},
            country={China}}

            
\affiliation[second]{organization={Guangzhou Institute of Geography, Guangdong Academy of Sciences}, addressline={100 Xianlie Middle Road}, city={Guangzhou}, postcode={510070}, state={Guangdong}, country={China}}

\begin{abstract}

Point cloud-based understanding has become an important enabler for facility operation and maintenance involving indoor building components. However, existing methods output only discrete labels without explaining component functions or natural language interactions. This paper proposes Building-MLLM, a point cloud-centered multimodal large language model (MLLM) for indoor components, which models point clouds and instructions to generate responses across Simple Recognition, Complex Captioning, and Multi-Engineering Question Answering tasks. Building-MLLM addresses semantic concentration through four domain-specific mechanisms: Point Information Enhancer for task-relevant semantics, Geometry-Preserving Regularization preventing geometric erosion, fixed textual prefix for domain stabilization, and multi-dimensional LoRA balancing recognition with reasoning. A multi-constraint progressive instruction-generation engine is developed to compile a synthetic point cloud-text dataset with 4,198 objects, 37,782 instruction-following pairs, and 47 categories. Experiments show that Building-MLLM achieves 88.00\%, 65.10\%, and 68.14\% on the three task types, respectively, demonstrating superior indoor component language understanding and providing initial generalisability in transfer inference on other real-world datasets.

\end{abstract}





\end{frontmatter}


\section{Introduction}
\label{introduction}

The architecture, engineering, and construction (AEC) industry faces mounting pressure to modernize facility operation and maintenance (O\&M) practices in response to aging infrastructure, increasing building complexity, and tightening sustainability regulations \citep{li2025integrating, yue2024deep}. Traditional maintenance workflows, which rely on manual inspection, paper-based documentation, and reactive repair strategies, are ill-suited for modern buildings that integrate hundreds of mechanical, electrical, and plumbing (MEP) components across multiple interconnected systems \citep{yin2023label}. These limitations manifest as inefficient resource allocation, prolonged downtime during failure diagnosis, and difficulty maintaining as-built documentation accuracy over a building's lifecycle \citep{zhang2026semi}. To address these challenges, the industry is increasingly adopting digital twin technologies and Building Information Modeling (BIM)–driven Computerized Maintenance Management Systems (CMMS) that promise proactive, data-driven facility intelligence \citep{ersoz2026evaluating}. By providing accurate geometric representations of indoor building components, 3D point clouds have gradually become an important medium linking physical representations with intelligent facility management \citep{liu2025digital}. Automatically acquiring semantic information from component-level point clouds, where building components serve as the smallest operational and analytical units, supports intelligent facility system development, analysis, and maintenance \citep{yue2026advancements}. However, most existing point cloud-based studies on building components primarily focus on component category recognition or classification \citep{yin2022automated,teclaw2025contextnet}, whose outputs are usually limited to discrete class labels and therefore cannot provide engineers with deeper engineering understanding or decision-making support. The vision of achieving automatic, language-based understanding of indoor building components from point cloud data, such as generating detailed component descriptions, explaining functional roles, or providing guidance, aims to bridge the semantic gap between geometric perception and language understanding. Realizing this capability would facilitate language-centered human–computer interaction, knowledge querying, and intelligent decision-making workflows, thereby enhancing the practical value of point cloud technologies in intelligent facility management.

Recent advances in large language models (LLMs) have catalyzed the emergence of multimodal LLMs (MLLMs), enabling unified reasoning and instruction following across visual and linguistic inputs \citep{liu2023visual,achiam2023gpt}.  These models offer a promising paradigm for engineering-scale understanding, as demonstrated in applications such as landslide analysis \citep{areerob2025multimodal}, crack detection \citep{wang2025crack}, and construction-site monitoring \citep{wong2025enhancing} where image-conditioned MLLMs generate diagnostic insights and actionable recommendations superior to traditional recognition models. Nevertheless, indoor building components exhibit domain-specific complexity, including fine-grained geometry, diverse topology, high intra-class variation, and function-oriented semantics, which cannot be adequately captured by 2D imagery alone \citep{armeni20163d}. In contrast, point clouds provide complete 3D geometry and explicit spatial relationships, making them an indispensable modality for deep semantic understanding at the component level \citep{yin2021automated}. While point cloud MLLMs are still in an early stage, several pioneering studies have laid groundwork for this direction. Cap3D \citep{luo2023scalable} provides the first large-scale point cloud–text dataset for generic multimodal alignment; PointLLM \citep{xu2024pointllm} demonstrates point-cloud-conditioned object recognition and captioning; MiniGPT-3D \citep{tang2024minigpt} explores lightweight point cloud–language backbone; and GPT4Point \citep{qi2024gpt4point} and ShapeLLM \citep{qi2024shapellm} extend interaction diversity and spatial reasoning. Despite these advances, directly applying such general models to indoor building components remains challenging. The domain lacks high-quality, structured point cloud-text instruction-following dataset capable of supporting fine-grained recognition, semantic explanation, engineering reasoning, and interactive understanding. Moreover, alignment strategies developed for broad, diverse object spaces fail to produce task-suitable geometric-semantic representations in concentrated engineering domains. Finally, parameter-efficient tuning methods such as LoRA \citep{hu2022lora} have not been systematically investigated for balancing recognition, captioning, and multi-step reasoning within point cloud MLLMs.

To address the systemic gaps in engineering semantic understanding and language reasoning for indoor building component point clouds, and to promote a transition from geometry-based labels to language-oriented component understanding and engineering reasoning, this study develops a unified MLLM framework for indoor building components, termed Building-MLLM, which takes component-level point clouds as the core perceptual input and integrates natural language instructions to enable unified modeling and responses across Simple Recognition, Complex Captioning, and Multi-Engineering Question Answering (QA) tasks, with the goal of establishing a stable and extensible semantic bridge between geometric perception and language understanding. Around this framework, we systematically investigate model domain-adaptation design and training strategies, engineering-oriented data generation pipelines and dataset construction, as well as task-aligned evaluation methodologies, forming a relatively complete research and validation chain. This study provides richer and more multidimensional empirical evidence, enabling engineering practitioners to obtain diverse language-based outputs from point cloud data beyond traditional classification results, thereby more effectively supporting understanding, decision-making, and application within intelligent facility management systems.  The contributions of this work are summarized as follows:

\begin{itemize}
\item \textbf{Domain-Specific Point Cloud MLLM for Indoor Building Components.} We propose Building-MLLM, a point cloud–centered MLLM specifically designed for indoor building component-level understanding. Within a unified modeling paradigm, it simultaneously supports Simple Recognition, Complex Captioning, and Multi-Engineering QA (Knowledge Capability, Commonsense Reasoning, Advanced Engineering Reasoning, Functional Comparison, Constraint Reasoning, Spatial Relationship, and Embodied Interaction) tasks, spanning five building systems with fine-grained geometric discrimination.

\item \textbf{Geometry-Preserving Cross-Modal Alignment Strategy.} We introduce a triple-constraint alignment mechanism to mitigate semantic concentration in professional domains, where frozen point encoders in MLLMs fail to capture fine-grained geometric variations. Specifically, the Point Information Enhancer (PIE) refines task-relevant latent geometric representations through recursive fusion of multi-source semantic cues, Geometry-Preserving Regularization (GPR) is applied by partially unfreezing shallow LLM layers to prevent geometric erosion, and a fixed textual prefix provides domain guidance to balance higher-level semantic abstraction.

\item \textbf{Multi-Dimensional LoRA Optimization for Multi-Task Balance.}  We design a systematic parameter-efficient fine-tuning strategy that simultaneously optimizes four LoRA configuration dimensions (layer range, target modules, rank, scaling strategies) to achieve stable performance across heterogeneous tasks, balancing short-range geometric recognition with long-range semantic reasoning and constraint interpretation in mixed point-text sequences under limited-data settings.

\item \textbf{Progressive Instruction-Following Generation Engine and Synthetic Dataset.} We develop a three-stage data generation engine that compiles the first point cloud-text instruction-following dataset for indoor building components, comprising 4,198 objects across 47 categories and 37,782 instruction-following pairs. The engine systematically integrates ontology- and category-level constraints, cross-modal consistency filtering constructed with BLIP-2/CLIP/Shap-E \citep{radford2021learning,li2023blip,jun2023shap}, and GPT-4V \citep{achiam2023gpt} equipped with task-driven prompts, thereby enabling the establishment of a multi-constrained and coherent engineering semantic chain from perception to reasoning.

\item \textbf{Comprehensive Evaluation Framework with GPT-4 as Semantic Evaluator.} Beyond automatic metrics (BLEU, ROUGE, METEOR, Sentence-BERT, SimCSE), we establish task-specific GPT-4 evaluation prompts that assess semantic equivalence, engineering logic consistency, and factual accuracy, providing reproducible and verifiable evaluation for multimodal generation in professional domains where surface-level lexical matching fails to capture correctness.

\end{itemize}

This work is positioned not as a complete O\&M knowledge agent, but as a component-level geometry-language understanding module. By moving beyond discrete label prediction toward richer semantic understanding, it provides an informative semantic entry point for linking point cloud perception with maintenance knowledge systems, and future embodied maintenance workflows. 

The remainder of this study is organized as follows. Section \ref{related_work} reviews prior research on point cloud–based understanding of indoor building components and outlines the key challenges in applying point cloud MLLMs to this domain. Section \ref{methoddataset} describes the proposed instruction-following data generation engine and the constructed dataset. Section \ref{methodology} presents Building-MLLM framework, detailing its domain-specific architectural design and training strategies. Section \ref{experiments} details the experimental setup, evaluation metrics, and qualitative and quantitative results, as well as two case-based analyses, including GPT-4–based evaluation cases and real-world dataset inference studies. Finally, Section \ref{conclusion} concludes the paper and highlights future research directions.

\section{Related Work}
\label{related_work}

 This section systematically reviews prior studies on indoor point cloud–based component recognition in building indoor and discusses the relevant background underpinning our research. Specifically, it covers the semantic gap in indoor point cloud component understanding, domain adaptation challenges of point cloud MLLMs, parameter-efficient fine-tuning strategies, and evaluation methodologies for multi-task response generation.

\subsection{From Geometric Labels to Language Understanding: The Semantic Gap}

In indoor environments, building components serve as the fundamental units for operation and analysis, and their geometric characteristics with semantic categories form a critical foundation for intelligent modeling, digital twin construction, and facility management systems \citep{mirzaei20223d,yue2026point}. Automatically acquiring component-level semantic information from indoor point clouds and effectively associating geometric shapes with semantic categories not only supports the automated generation and updating of 3D information models, but also enables asset management, operation and maintenance analysis, and intelligent decision-making \citep{li2026vision,yue2026enhancing}. Consequently, component-level classification and recognition from indoor point clouds have become important research directions in intelligent facility management.

With the rapid development of computer vision and point cloud technologies, indoor component recognition have progressively evolved from traditional geometry-driven and handcrafted feature approaches toward data-driven paradigms centered on deep learning \citep{yue2026transfer}. A typical workflow begins with an encoder, including PointNet-based MLP variants \citep{qi2017pointnet,qi2017pointnet++} and Transformer-based architectures \citep{zhao2021point}, to extract local and global geometric features, followed by a decoder that generates object-level semantic labels. For example, Koo et al. \citep{koo2021automatic} employed PointNet to achieve fine-grained classification of typical indoor components, such as walls, doors, and windows. Emunds et al. \citep{emunds2022sparse} proposed a sparse convolution–based network that captures geometric-semantic characteristics of indoor components, significantly improving inference efficiency while maintaining competitive performance. In addition, Teclaw et al. \citep{teclaw2025contextnet} proposed ContextNet, which incorporates contextual information from neighboring components to improve classification accuracy by modeling spatial relationships. Focusing on pipe components, Yin et al. \citep{yin2022automated} introduced a squeeze-and-excitation mechanism and proposed the SE-PseudoGrid to enhance classification robustness under complex conditions. For general datasets containing diverse component types, including furniture \citep{uy2019revisiting}, Wu et al. \citep{wu2024point} demonstrated strong classification adaptability through lightweight network designs and architectural refinements. Overall, these component-level point cloud classification studies differ from scene-level semantic segmentation methods \citep{zhai2022bim,yue2026weakly} that rely on local inter-component boundaries, and have achieved notable progress in object-level shape modeling, robustness, and cross-domain adaptability.

Despite these advancements, existing methods remain confined to label-space semantics: they output category names but cannot explain \textit{why} a component is classified as such, \textit{where} it should be installed, \textit{how} it differs from similar components, or \textit{what} diverse attributes are involved. This limitation stems from a fundamental paradigm constraint—current pipelines treat semantics as discrete labels rather than as compositional language capable of encoding geometric attributes, functional properties, spatial relationships, and operational constraints. These limitations leave meaningful scope for our work to explore the integration of indoor component point clouds with LLMs, by treating point cloud geometry and natural language as joint inputs to enable language-grounded understanding of building components. By moving beyond label-space semantics toward compositional, language-based representations, this paradigm supports more expressive, interpretable, and semantically rich understanding of components, laying the foundation for advanced intelligent facility management.

\subsection{Point Cloud MLLMs: The Domain Adaptation Challenge}

In recent years, point cloud–centric MLLMs have emerged, offering a promising shift for 3D understanding tasks from traditional label prediction paradigms toward language generation and semantic reasoning \citep{hong20233d}. Architecturally, these methods inherit the design principles of visual MLLMs \citep{liu2023visual} but replace the 2D image modality with 3D point clouds. Specifically, a pipeline composed of a point cloud encoder, a projection module, and a LLM is employed to map and align point cloud features into the linguistic semantic space, thereby enabling unified language understanding and expression of 3D objects by seeing point clouds and generating language descriptions.

Among representative studies, PointLLM \citep{xu2024pointllm} was the first to systematically explore the feasibility of point cloud–language alignment. It employs a frozen Point-BERT encoder \citep{yu2022point} for point cloud representation and enables language description generation for 3D objects through feature projection and full-parameter training of the LLM backbone, achieving superior performance over 2D-based baseline \citep{dai2023instructblip} across several tasks. Subsequently, ShapeLLM \citep{qi2024shapellm} introduced a multi-view image distillation mechanism to compensate geometric perception from point clouds and further enhanced 3D understanding through instruction optimization. GPT4Point \citep{qi2024gpt4point} went a step further by constructing a unified point cloud–language representation space and exploring the extension of language model capabilities toward 3D generation tasks. In contrast, MiniGPT-3D \citep{tang2024minigpt} leverages semantic priors from existing visual MLLMs as a bridging mechanism and significantly reduces training costs for point cloud MLLMs through a multi-stage alignment strategy. At the data level, large-scale training of point cloud MLLMs is commonly supported by the general-purpose Objaverse dataset \citep{deitke2023objaverse}, which contains over 800,000 3D objects. By integrating BLIP-2 \citep{li2023blip}, CLIP \citep{radford2021learning}, and GPT-4 \citep{achiam2023gpt}, point cloud–text pairs are constructed. Specifically, the Cap3D dataset \citep{luo2023scalable} generates candidate captions using BLIP-2, performs semantic matching via CLIP, and aggregates captions with GPT-4, resulting in approximately 785,000 point–text pairs. Building upon this foundation, subsequent studies \citep{xu2024pointllm} further employ GPT-4 to expand simple captions into more compositional and descriptive forms, providing richer supervision signals for instruction learning and capability scaling in point cloud MLLMs.

However, direct transfer of these general-domain models to professional engineering scenarios faces two critical challenges. First, the semantic concentration problem: indoor building components exhibit fine-grained geometric variations within concentrated categorical spaces (e.g., different valve types, duct fittings, pipe junctions) \citep{yin2022automated, utkucu2024classification}, whereas general models are optimized for coarse-grained discrimination across diverse object classes (furniture, vehicles, animals) \citep{uy2019revisiting,deitke2023objaverse}. Frozen point cloud encoders \citep{yu2022point,xue2023ulip}, while stable for broad domains, cannot adapt to domain-specific geometric cues required for professional discrimination. Second, the instruction diversity problem: existing instruction-following datasets and their generation pipelines emphasize shape description and spatial reasoning for general objects but lack engineering-oriented knowledge chains—including precise engineering ontology recognition, functional semantics, constraint interpretation, installation logic, and embodied operational procedures—that are essential for facility management applications. These gaps manifest as significant performance drops when state-of-the-art (SOTA) models are evaluated on indoor component datasets, as evidenced by our validation experiments. In response to these challenges, we aim to develop a domain-specific point cloud MLLM for indoor building components, in which multiple alignment mechanisms are introduced to enhance domain adaptation. 
In addition, we revisited the instruction generation process and, to address semantic overgeneralization, hallucination, and the difficulty of enriching semantic content in previous data generation pipelines, we proposed a progressive instruction design guided by ontology priors and multiple engineering constraints, spanning simple recognition, complex description, and engineering-oriented question answering. Through those designs, our goal is to improve the domain-aware multi-task learning and reasoning capabilities of MLLMs for indoor building components.

\subsection{Parameter-Efficient Cross-Modal Fine-Tuning: The Multi-Task Balance Problem}

Current point cloud MLLM training paradigms adopt a two-stage strategy: geometry-language alignment (freeze encoder, train projector) followed by instruction fine-tuning \citep{yin2024survey}. In the fine-tuning stage, mainstream approaches fully unfreeze the entire LLM backbone to enable instruction-following capabilities \citep{xu2024pointllm,qi2024shapellm}. While effective for general domains with abundant data, this full-parameter strategy faces critical limitations when applied to professional scenarios with limited annotated corpora and diverse task requirements. The resource constraint: facility management datasets are orders of magnitude smaller than general vision-language corpora (thousands vs. millions of samples), making full LLM fine-tuning prone to overfitting and requiring prohibitive computational resources. The task heterogeneity challenge: unlike single-task settings (e.g., pure captioning), engineering applications demand simultaneous competence in short-range geometric recognition (component identification), medium-range semantic generation (detailed descriptions), and long-range reasoning (multi-turn QA with constraint interpretation). Existing parameter-efficient methods like LoRA \citep{hu2022lora} have been extensively studied in pure-language and 2D vision-language \citep{liu2023visual} settings, but their effectiveness in mixed point-text sequences—where geometric and linguistic tokens are tightly coupled—remains underexplored. Specifically, the interplay between LoRA configuration dimensions (layer depth, module selection, rank capacity, scaling strategies) and multimodal task balance has not been systematically investigated. This gap motivates our multi-dimensional LoRA optimization strategy, which enables controllable adaptation to domain-specific distributions while maintaining stable performance across heterogeneous tasks.

\subsection{Evaluation Metrics for Response Generation: Multi-Task Evaluation Prompt Design}

Evaluating the quality of responses generated by LLMs and MLLMs requires a scientifically grounded and systematic framework \citep{sai2022survey}. Traditional automatic evaluation metrics—such as BLEU \citep{papineni2002bleu}, ROUGE \citep{lin2004rouge}, and METEOR \citep{banerjee2005meteor}—are widely used in machine translation, text generation, and image captioning tasks. These metrics measure explicit overlap between model outputs and reference texts through n-gram precision, recall, and lexical matching, thereby emphasizing surface-level similarity. In addition, embedding-based semantic metrics such as Sentence-BERT and SimCSE capture sentence-level semantic similarity \citep{reimers2019sentence, gao2021simcse}, offering more reliable assessments for long-form descriptions and narrative content. However, these methods remain limited in their ability to handle diverse yet semantically equivalent expressions, and they struggle to evaluate factual consistency and domain-specific engineering logic \citep{xu2024pointllm}.  With the rapid advancement of large models, GPT-4–based evaluators have emerged as a more reliable alternative, demonstrating near human-level performance in tasks such as captioning, summarization, and question answering \citep{xu2024pointllm,tang2024minigpt}. Engineering studies \citep{areerob2025multimodal} have further validated the domain reliability of the GPT-4 evaluator. Given that our study focuses on engineering-oriented, multi-task point–text generation, relying solely on traditional metrics is insufficient to comprehensively assess model capabilities. Therefore, constructing a set of GPT-4 evaluation prompts tailored to different task types is essential for achieving accurate and effective evaluation of multi-task responses.

\section{Construction of Indoor Building Component Instruction Dataset}
\label{methoddataset}

This section introduces the proposed Progressive Instruction-Following Generation Engine and further presents the first synthetic point cloud-text instruction-following dataset for indoor building components compiled using this engine.

\subsection{Progressive Instruction-Following Generation Engine}

\begin{figure}[!htb]
\centering
\includegraphics[width=\textwidth]{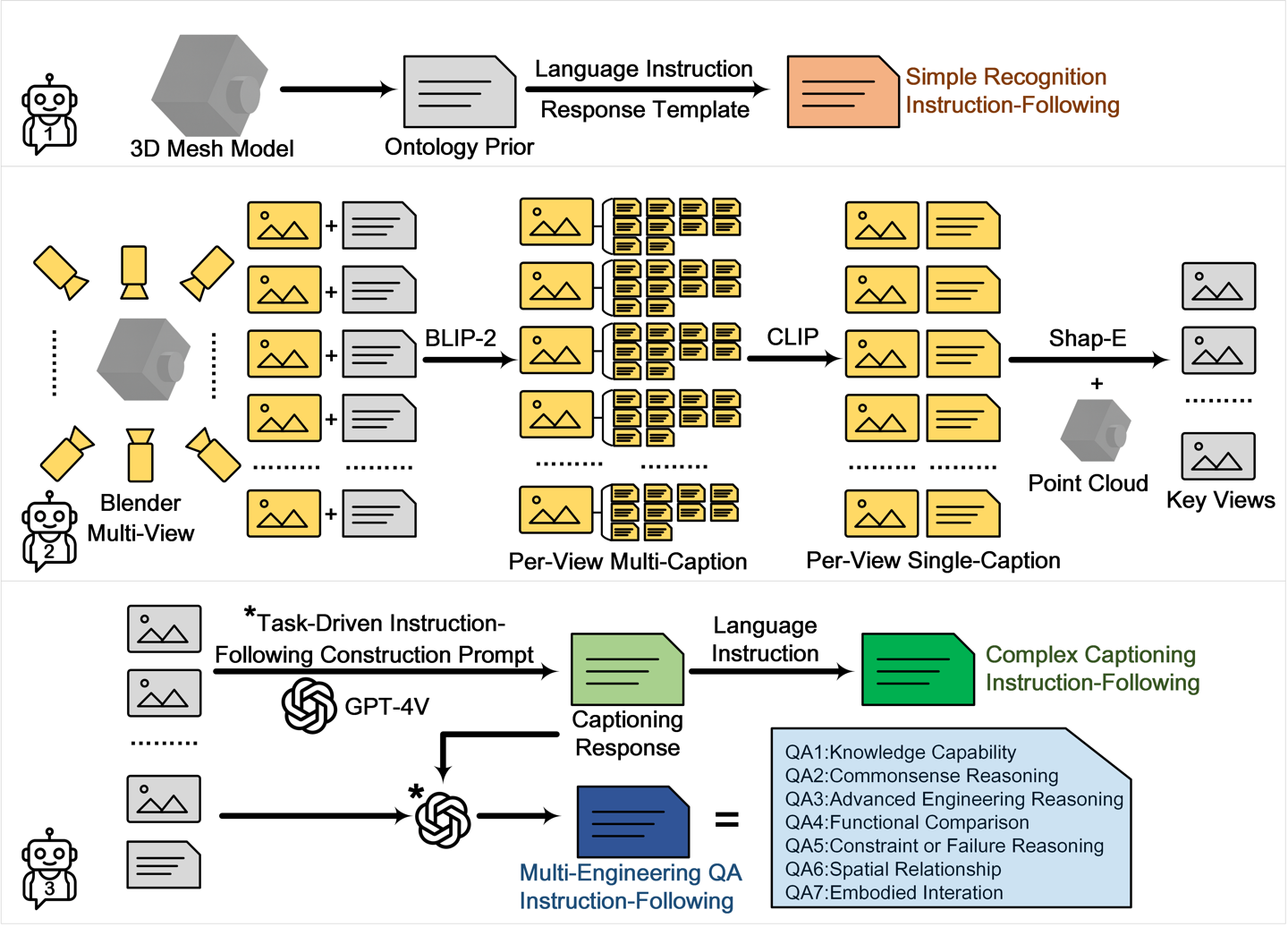}
\caption{Progressive instruction-following generation engine for constructing point cloud-text training data with semantic enrichment.}
\label{F2}
\end{figure}

To equip the model with diversified linguistic capabilities and robust comprehension of varied human instructions, the proposed engine, consisting of three steps and illustrated in \autoref{F2}, is designed to generate multi-task point cloud-text instruction-following spanning Simple Recognition, Complex Captioning, and Multi-Engineering QA. In this progressive pipeline, Step 1 establishes categorical alignment through ontology-constrained instruction and response templates; Step 2 selects semantically representative rendered views through cross-modal consistency filtering using BLIP-2, CLIP, and Shap-E, thereby reducing hallucinations at the source; and Step 3 uses GPT-4V with task-driven prompts to generate Complex Captioning and Multi-Engineering QA data, forming a knowledge chain from visual perception to engineering reasoning. The Multi-Engineering QA component further encompasses seven categories of tasks covering knowledge, reasoning, and interaction. Specifically, Knowledge Capability is defined as the ability to represent factual knowledge of indoor components; Commonsense Reasoning refers to reasoning based on general physical intuition and everyday experience; Advanced Engineering Reasoning captures reasoning grounded in engineering principles, including structural behavior, design logic, and performance analysis; Functional Comparison describes the ability to analyze differences between components in terms of function and usage; Constraint Reasoning focuses on reasoning under practical constraints, such as safety regulations, installation limitations, and environmental conditions; Spatial Relationship denotes the understanding of the position, orientation, and connectivity of components in 3D space; and Embodied Interaction characterizes the ability to reason about physical interactions with components, including installation, operation, and maintenance processes.

\newpage

\textbf{Step 1: Simple Recognition Text Instruction-Following Generation.} This step constructs concise instruction–response samples for each 3D object within the Simple Recognition task. We first design 12 natural-language instructions and 10 standardized response templates (\autoref{T1}). For each object, one instruction and one response template are randomly selected, and the object ontology prior category is injected into the template to form a unique instruction-following pair (e.g., "What is this?"—"This is a variable air volume.").Through the use of diverse instruction and response templates, this step enhances the model's robustness to instruction variation, while ontology-constrained responses ensure high semantic fidelity. These samples collectively provide a strong basis for semantic alignment and instruction tuning in point cloud MLLMs.

\begin{table*}[!htb]
\footnotesize
\captionsetup{margin=0cm}
\caption{Designed natural language instructions and answer templates for the Simple Recognition task.}
\centering
\begin{tblr}{
  width=\textwidth,
  colspec={p{0.58\textwidth} p{0.36\textwidth}},
  hline{1-2,14} = {-}{},
}
\textbf{Question} & \textbf{Answer Template} \\

1. What is this? & 1. This is a \{\}. \\
2. Based on the shape, what do you think this point cloud is? & 2. 3D model of a \{\}. \\
3. What is the most likely object this point cloud represents? & 3. The object presents a \{\}. \\
4. Give a concise description of the object shown in this 3D point cloud. & 4. A \{\}. \\
5. Which category does this 3D object most likely belong to? & 5. This is a \{\} point cloud. \\
6. Describe the 3D point cloud object briefly. & 6. The shape corresponds to a \{\}. \\
7. Deliver a quick description of the object represented here. & 7. You are looking at a \{\}. \\
8. Summarize the object rendered by this sparse point cloud briefly. & 8. This seems to be a \{\}. \\
9. Identify the object shown from the point cloud. & 9. This object belongs to the \{\}. \\
10. Provide a simple description for this collection of 3D points. & 10. This is a geometric object of a \{\}. \\
11. How would you interpret this 3D point cloud? & ~ \\
12. Can you tell what structure is formed by these 3D points? & ~ \\

\end{tblr}
\label{T1}
\end{table*}

\textbf{Step 2: Key Views Selection.} This step aims to select semantically representative rendered key views for each 3D object, providing high-quality visual references for subsequent instruction-following generation. We first visualize each 3D model using a Blender–Phong rendering pipeline, placing the camera uniformly at 12 viewpoints around the object and generating 512×512 images with EEVEE to maximize geometric detail coverage. BLIP-2 is then used to produce textual descriptions for each view, incorporating category priors during generation and expanding each view into 10 COCO-style candidate captions via nucleus sampling. CLIP computes the image–text semantic similarity to filter out inconsistent or noisy captions, retaining the most aligned image–text pair for each view. Conventional approaches often aggregate these single-view descriptions using GPT-4 to produce a global 3D caption. Such aggregation can introduce discrepancies between visual and linguistic cues, amplify hallucinations originating from individual views, and confine the outputs to caption-level representations, thereby limiting their extensibility. To mitigate these issues, we avoid direct aggregation and instead use the selected image–text pairs for cross-modal consistency evaluation. Specifically, we employ Shap-E, a text-to-3D diffusion model, to compute reconstruction errors and consistency scores between each rendered view–text pair and the latent point cloud representation. Views are then ranked using these scores, and the top six with the highest scores are selected, as they best reflect the object's geometric and semantic characteristics. This number is determined through repeated experiments across different categories to achieve an optimal balance. This cross-modal alignment mechanism effectively suppresses hallucinations at the source. By integrating BLIP-2, CLIP, and Shap-E into an automated, fully open-source filtering pipeline, our method adaptively identifies the most semantically informative views, substantially improving the accuracy of subsequent GPT-4V instruction generation while markedly reducing the API cost of large-scale dataset construction.

\begin{table}[!htb]
\footnotesize
\captionsetup{margin=0cm}
\caption{Task-Driven Instruction-Following Construction Prompt designed for Complex Captioning and Multi-Engineering QA point cloud text instruction following.}
\begin{tblr}{
  hlines,
  colspec={Q[1.5cm] X[15cm]},
  cell{2}{2} = {valign=t},
}

\textbf{System Prompt} &
You are a multi-category 3D computer visual assistant for interior building environments, with professional visual analysis capabilities and domain knowledge of mechanical, electrical, plumbing, furnishing, structural, and architectural systems.
You must provide accurate, structured, and objective responses based solely on rendered images of labeled 3D models.
Always follow the requested output format strictly.
Avoid subjective descriptions, vague terms, and dialogue simulation.
Your responses should reflect expert-level understanding of geometry, structure, function, and spatial relationships, and related aspects. \\

\textbf{Input} &
object\_id, object\_class, and key multi-view images \\

\textbf{User Prompt} &
Url \{key multi-view images\}: You are given a set of multi-angle rendered images of a 3D object labeled as \{object\_class\}.
Based on the image contents, first complete Task 0 as a detailed visual analysis. Then, use the image contents and your Task 0 result as factual context to answer Tasks 1–7. Return the result in the specified JSON format.

Task 0 – Complex Captioning: Generate a 50–60 word objective description of the object, focusing on category, structure, geometry, connection forms, and distinguishing features.

Task 1 – Knowledge Capability: Provide a factual QA focused on \{task1\}. ($\leq$20 words) 

Task 2 – Commonsense Reasoning: A QA about \{task2\}. ($\leq$20 words)  

Task 3 – Advanced Engineering Reasoning: A QA about \{task3\}. ($\leq$20 words)  

Task 4 – Functional Comparison: A QA about \{task4\}. ($\leq$20 words)  

Task 5 – Constraint Reasoning: A QA about \{task5\}. ($\leq$20 words)  

Task 6 – Spatial Relationship: A QA about \{task6\}. ($\leq$20 words)  

Task 7 – Embodied Interaction: List 3–5 steps for \{task7\}, each step $\leq$12 words.

\textbf{Output Format:} \{

  "object\_id": "\{object\_id\}",  
  
  "object\_class": "\{object\_class\}",  
  
  "tasks": \{
  "Complex\_Captioning": "...",  
    
    "Knowledge\_Capability": {"question": "...", "answer": "..."},  
    
    "Commonsense\_Reasoning": {"question": "...", "answer": "..."},  
    
    "Advanced\_Engineering\_Reasoning": {"question": "...", "answer": "..."},  
    
    "Functional\_Comparison": {"question": "...", "answer": "..."},  
    
    "Constraint\_Reasoning": {"question": "...", "answer": "..."},  
    
    "Spatial\_Relationship": {"question": "...", "answer": "..."},  
    
    "Embodied\_Interaction": {"question": "...", "answer": "1. 2. 3. 4. 5."}  
    \}  
\} \\

\textbf{Stochastic Sampling} &
task1 = random.choice(["its category definition", "the object's function", "a real-world usage scenario"])  

task2 = random.choice(["its physical traits in daily use", "user interaction features", "visible working status cues", "comfort-related attributes"])  

task3 = random.choice(["structural principles", "engineering design logic", "efficiency or safety principles", "heat, flow, or signal behavior"])  

task4 = random.choice(["functionality difference across similar components", "structural contrast with similar components", "usage scenario comparison"])  

task5 = random.choice(["usage limitations or restrictions", "common installation errors", "regulatory or safety rules", "environmental constraints", "user behavior to avoid"])  

task6 = random.choice(["typical position in a room or system", "orientation relative to walls or fixtures", "connection layout with other components", "installation-facing direction"])  

task7 = random.choice(["daily use", "installation steps", "routine maintenance"]) \\
\end{tblr}
\label{T2}
\end{table}

\textbf{Step 3: Complex Captioning and Multi-Engineering QA Text Instruction-Following Generation.} Interactive multimodal generation with GPT-4V enables complex, coherent descriptions and naturally supports semantic extensions. To leverage this capability, we design a Task-Driven Instruction-Following Construction Prompt (\autoref{T2}) using API, Within this, the System Prompt specifies generation rules, roles, and stylistic constraints to ensure consistency and objectivity, while the User Prompt injects dynamic, task-specific instructions to achieve flexible adaptation across different objects and scenarios.The generation process begins by utilizing key views and ontology priors of the object to produce the Complex Captioning response—a detailed, multi-attribute description that forms a comprehensive semantic representation of the object. Building upon this caption, the system further utilizes the key views to automatically construct a seven-round Multi-Engineering QA sequence covering Knowledge Capability, Commonsense Reasoning, Advanced Engineering Reasoning, Functional Comparison, Constraint Reasoning, Spatial Relationship, and Embodied Interaction. To introduce controlled randomness without compromising the semantic boundaries of these seven tasks, a stochastic sampling mechanism is applied to task-specific keyword prompts. This approach mitigates model bias toward fixed candidates while enhancing the diversity and robustness of the generated samples. Each object's Complex Captioning response is supplemented with an instruction randomly selected from the templates designed in \autoref{T3}. Together, these processes construct the Complex Captioning and Multi-Engineering QA instruction-following data.

\begin{table}[!htb]
\footnotesize
\captionsetup{margin=0cm}
\caption{Designed natural language instructions for the Complex Captioning task.}
\centering
\begin{tblr}{
  colspec={Q[l,wd=0.47\textwidth] Q[l,wd=0.47\textwidth]},
  hline{1-2,8} = {-}{},
}
\textbf{Question} &  \textbf{Question} \\
1. Can you tell me more about this?
& 7. Could you delve deeper into this? \\
2. I'm curious about this, could you clarify it for me?
& 8. I want to know more about this, can you help? \\
3. Please detail the specific features of this point cloud.
& 9. Can you walk me through the details of this object? \\
4. Would you mind offering additional information on this?
& 10. Can you provide a comprehensive account of this object? \\
5. Could you provide more info about this?
& 11. Elaborate on the details of this point cloud, please. \\
6. Could you assist me in understanding this better?
& 12. Kindly offer a precise and structured explanation of what these 3D points indicate. \\
\end{tblr}
\label{T3}
\end{table}

These instruction-following data are ultimately stored in a structured JSON format shown in \autoref{T4}, this schema represents the content of each point cloud instance under specific tasks.

\begin{table}[!htb]
\footnotesize
\centering
\caption{Instruction-Following schema.}

\begin{tblr}{
  colspec = {l l l l l},
  cell{3}{1} = {r=4}{},
  hline{1-3,7} = {-}{},
}
\textbf{Object\_id} & \SetCell[c=4]{c}{} & & & \\ 
\textbf{Task\_type} & & Simple Recognition & Complex Captioning & \SetCell[c=2]{c}{Multi-engineering QA} & \\
\textbf{Conversations} & USER & point\textbackslash{}n\{Instruction\} & point\textbackslash{}n\{Instruction\} & point\textbackslash{}n\{Instruction 1\} \\
              & ASSISTANT & \{Response\}/s & \{Response\}/s & \{Response 1\}/s \\
              & USER &  & &  \{Instruction ..\} \\
              & ASSISTANT &  & & \{Response ..\}/s \\
\end{tblr}
\label{T4}
\end{table}

\subsection{Dataset Statistics and Properties}

During dataset construction, we incorporated five fundamental indoor building systems including Structural and Architectural, Furnishing, Electrical, Mechanical, and Plumbing. These systems collectively define the semantic foundation and operational structure of indoor environments, encompassing the most representative categories that collaboratively sustain a multidimensional balance among structural integrity, environmental control, energy management, human comfort, and safety assurance.

We collected 3D mesh models from several public repositories, including IFCNetCore \citep{emunds2021ifcnet}, BIM Object \citep{bimobject2023}, BIM Store \citep{bimstore2023}, NBS Source \citep{nbsenterprises2023}, the CIC BIM Object Library \citep{cic2023bimobjectlibrary}, and BIMGEOM \citep{collins2021bimgeom}. After deduplication and quality checks \citep{utkucu2024classification}, all meshes were converted into point clouds through area-weighted uniform surface sampling, which allocates points proportionally to face areas to ensure balanced geometric coverage. Each object was sampled with 12,000 points and represented by three-dimensional coordinates (x, y, z) with randomly assigned colors (r, g, b). In parallel, our proposed engine generated instruction-following data for each object, establishing a one-to-one mapping between point clouds and their corresponding text data using model IDs, ultimately yielding a point cloud-text instruction-following dataset tailored for indoor building components.

\begin{figure}[!htb]
\centering
\includegraphics[width=\textwidth]{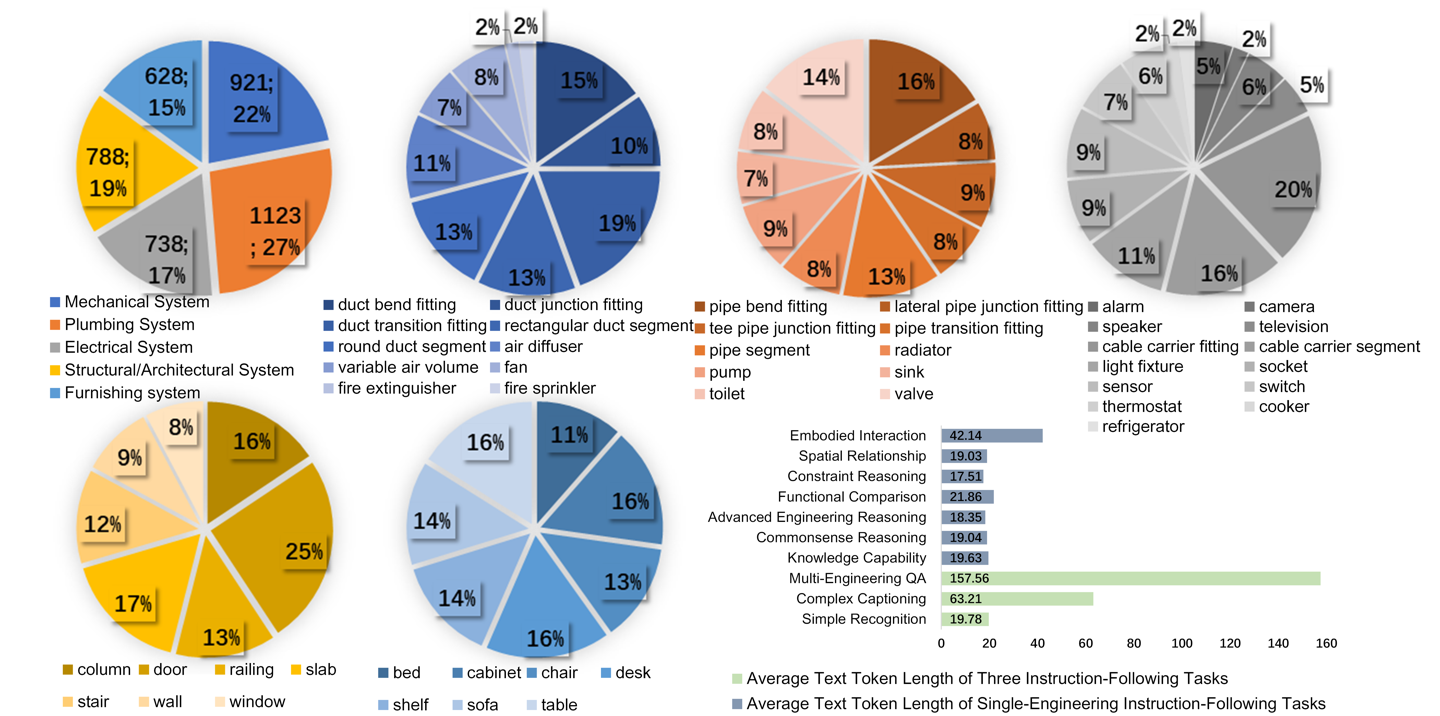}
\caption{Object category and instruction-following statistics in the proposed dataset.}
\label{F1}
\end{figure}

The dataset consists of 4,198 point clouds covering 47 component categories across five systems, together with one-to-one instruction-following data for three tasks—Simple Recognition, Complex Captioning, and Multi-Engineering QA—resulting in a total of 37,782 instruction-following pairs. As shown in \autoref{F1}, the proportion of components in each system ranges from 15\% to 27\%, indicating a relatively balanced distribution. Each system contains 7 to 15 categories, and all category
names follow the labels of the original source models. We used GPT2TokenizerFast to compute the token lengths of the textual instructions. For each point cloud, the average token counts of the three tasks are 19.78, 63.21, and 157.56, respectively, reflecting the increasing semantic complexity from recognition to detailed captioning and multi-round engineering interaction. Within the Multi-Engineering QA task, the token lengths of different engineering sub-tasks are comparable, whereas the Embodied Interaction task exhibits noticeably longer outputs due to its multi-step operational descriptions.

\begin{figure}[!htb]
\centering
\includegraphics[width=\textwidth]{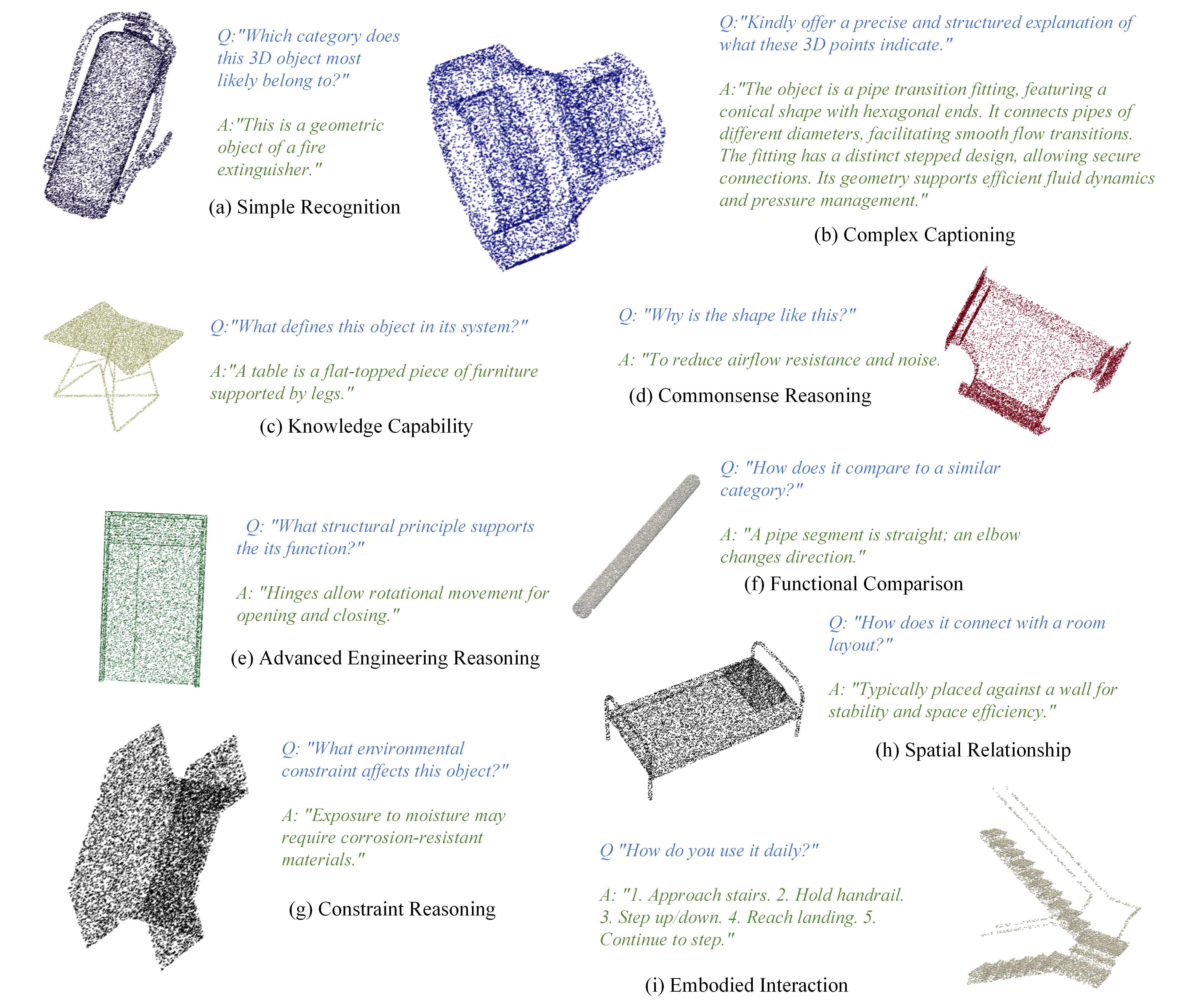}
\caption{Examples of point cloud-text instruction-following data for different tasks.}
\label{F3}
\end{figure}

Some examples from the constructed dataset are shown in \autoref{F3}, the Simple Recognition task focuses on categorical identification; the Complex Captioning task extends this to fine-grained descriptions of multiple attributes, structural characteristics, and functional properties; and the Multi-Engineering QA task further incorporates higher-level semantic understanding and engineering-oriented reasoning. Accordingly, this dataset is appropriately understood as a form of expert-informed and constraint-guided synthetic supervision.

To check and assess the quality of the dataset, we randomly sampled approximately 20\% of the objects from each category, resulting in a subset of 800 point clouds and their corresponding texts for domain-specific manual review. The review was conducted by a team of four doctoral researchers in civil engineering, who examined the semantic alignment and content accuracy between the generated texts and the corresponding point clouds. 

For the Simple Recognition task, the generated texts were deterministically produced based on category priors and predefined templates, without involving free-form language generation. Therefore, this task did not introduce generative uncertainty and was not treated as the primary focus of manual quality inspection. For the Complex Captioning task, the reviewers mainly checked whether the generated descriptions contained missing attributes, incorrect statements, or semantically incomplete content. The evaluation criteria were guided by the prompt design, with particular attention to the geometric characteristics, functional properties, and structural relationships of the point cloud objects. For the Multi-Engineering QA task, the reviewers evaluated each response by jointly considering the point cloud object and the corresponding question. The assessment focused on whether the generated answers were consistent in terms of engineering semantics, spatial relationships, and operational logic, thereby ensuring that the content was reasonable, interpretable, and engineering-feasible under the task constraints.

Specifically, reviewers judged whether each automatically generated text required semantic correction based on task-specific validity. During the manual review process, any generated text showing semantic ambiguity, incompleteness, missing key information, or logical inconsistency was strictly marked as ``Need correction''. Accordingly, no intermediate state such as ``Partially correction'' was defined in the evaluation protocol. Even if a generated text was generally reasonable, it was still classified as ``Need correction'' once any deficiency or deviation was identified in key attributes, functional description, or engineering semantics. This judgment was formulated as a binary decision, i.e., ``No correction'' or ``Need correction''. To quantify the consistency of reviewers on this binary decision, we employed Cohen's Kappa coefficient. \autoref{tab:kappa_matrix} presents the 2$\times$2 agreement matrix between two reviewer groups (Group~1 and Group~2), each consisting of two independent reviewers. Here, $A$ and $D$ denote the number of samples for which both groups agreed that no correction or correction was needed, respectively, while $B$ and $C$ represent disagreement cases.

\begin{table}[!htb]
\centering
\caption{2$\times$2 agreement matrix used for Cohen's Kappa computation.}
\label{tab:kappa_matrix}

\begin{tabular}{c|cc}
\hline
 & Group 2 (No correction) & Group 2 (Need correction) \\
\hline
Group 1 (No correction) & $A = 748$ & $B = 5$ \\
Group 1 (Need correction) & $C = 6$ & $D = 41$ \\
\hline
\end{tabular}

\end{table}

Cohen's Kappa coefficient is defined as:
\begin{equation}
\kappa = \frac{p_o - p_e}{1 - p_e},
\end{equation}
where $p_o$ denotes the observed agreement rate and $p_e$ denotes the expected agreement by chance. The observed agreement rate is computed as:
\begin{equation}
p_o = \frac{A + D}{N},
\end{equation}
where $N$ is the total number of reviewed samples. The expected agreement rate by chance is estimated from the marginal distributions as:
\begin{equation}
p_e =
\left( \frac{A + B}{N} \cdot \frac{A + C}{N} \right)
+
\left( \frac{C + D}{N} \cdot \frac{B + D}{N} \right).
\end{equation}

Substituting the values in \autoref{tab:kappa_matrix}, we obtain
$p_o = 0.98625$ and $p_e = 0.89048$, resulting in a Cohen's Kappa coefficient of $\kappa = 0.874$.
This result indicates near-perfect agreement among reviewers on whether automatically generated texts require semantic correction, with very few disagreement cases. Moreover, the majority of samples were consistently judged as requiring no correction (94.125\% by Group~1 and 94.25\% by Group~2), while only 5.125\% of samples were jointly identified as requiring correction. These findings suggest that the automatically generated texts exhibit high initial semantic correctness and professional validity. Based on the review results, anomalous annotations were manually corrected, and the refined subset was subsequently used for model inference evaluation and test.

\section{Methodology}
\label{methodology}

Our proposed Building-MLLM addresses the identified gaps through a three-pronged strategy centered on domain-specific alignment and parameter-efficient adaptation. (1) At the data level, we construct a complete instruction-following pipeline that progressively enriches semantic granularity—from categorical labels to compositional descriptions to engineering reasoning chains—ensuring training signals cover the full knowledge spectrum required for facility intelligence (detailed in Section \ref{methoddataset}). (2) At the architectural level, we recognize that generic cross-modal alignment mechanisms fail in concentrated professional domains where subtle geometric differences carry critical semantic meaning. Therefore, we introduce a triple-constraint alignment strategy: a PIE that injects task-relevant geometric semantics through recursive feature learning, GPR that constrains shallow LLM layers to prevent geometric erosion during semantic updates, and a fixed textual prefix that stabilizes domain-specific linguistic patterns. (3) At the training level, we observe that multimodal instruction-following demands balancing short-range geometric grounding (recognition) with long-range semantic reasoning (captioning and QA), which cannot be achieved through uniform parameter updates. We therefore adopt a multi-dimensional LoRA optimization strategy that systematically controls layer-wise influence, module-wise capacity, rank-wise expressiveness, and scaling-wise magnitude to achieve stable multi-task performance under resource constraints. This design philosophy—domain-grounded data, geometry-preserving alignment, and task-balanced fine-tuning—forms the conceptual foundation of Building-MLLM and distinguishes it from general-domain multimodal frameworks.

\subsection{Revisiting baseline model: PointLLM}

\begin{figure}[!htb]
\centering
\includegraphics[width=\textwidth]{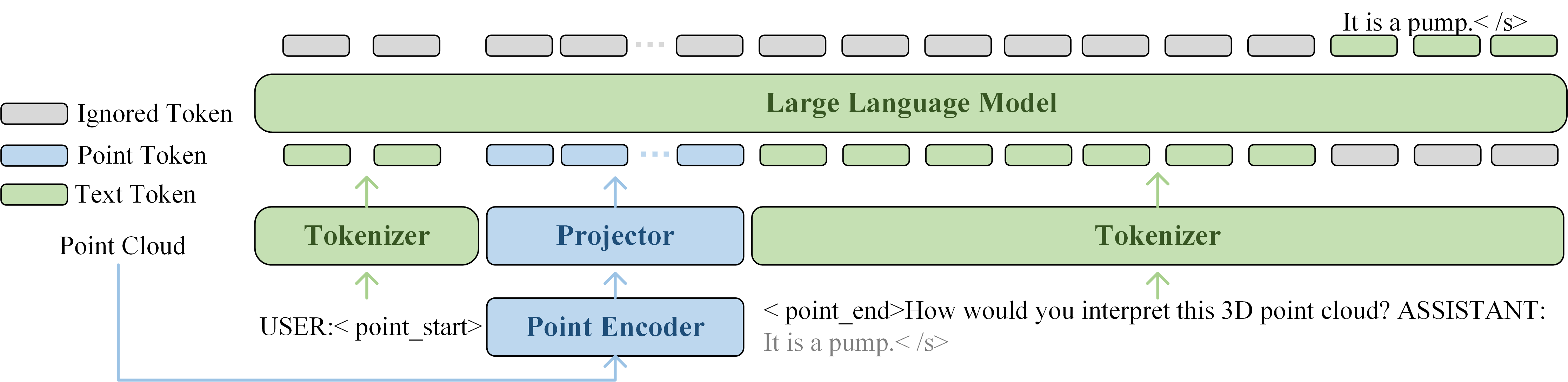}
\caption{Overview of PointLLM \citep{xu2024pointllm} architecture.}
\label{F4A}
\end{figure}

Our proposed Building-MLLM builds upon the foundational architecture of PointLLM, a state-of-the-art point cloud MLLM, while introducing domain-specific enhancements to address the unique challenges of indoor building component understanding. Before presenting our contributions, we briefly review the PointLLM framework to establish necessary context. PointLLM \citep{xu2024pointllm} (as shown in \autoref{F4A}) adopts a unified token-based architecture: textual inputs are tokenized and embedded into a high-dimensional language space, while point clouds are processed by a frozen geometric encoder (PointBERT) and projected into the same semantic space via learnable projection layers. These multimodal embeddings are concatenated into a unified sequence and fed into a pre-trained LLM (LLaMA-7B) for autoregressive generation. While effective for general-domain 3D objects, this design exhibits two critical limitations when applied to indoor building components: (1) the fixed-weight geometric encoder cannot adapt to fine-grained geometric variations in concentrated professional domains, and (2) uniform parameter updates during instruction fine-tuning fail to balance heterogeneous task requirements (recognition vs. reasoning). Our method addresses these limitations through the mechanisms detailed below. 

Building on this foundation, Building-MLLM adopts a two-stage training strategy for specialized point cloud–language understanding of indoor building components. In stage 1, we introduce the PIE module, the GPR mechanism, and a fixed textual prefix to enhance domain-relevant semantics and improve the stability of point cloud–language alignment. In stage 2, we design a multi-dimensional LoRA optimization strategy to accommodate diverse task requirements, including fine-grained geometric recognition, complex caption generation, and multi-turn engineering QA. Although instantiated on PointLLM in this study, the proposed modules are not inherently tied to this specific framework. Instead, they are designed around more general issues in point cloud–MLLM modeling, including frozen point cloud representations, point cloud–text mixed-sequence alignment, geometric information preservation, and heterogeneous task balancing. Therefore, as more advanced general-domain point cloud–MLLM frameworks become publicly available, the proposed framework can be transferred to other backbones with necessary task-driven adjustments according to specific engineering application requirements.

\subsection{Stage 1: Geometry-Preserving Cross-Modal Alignment}

\begin{figure}[!htb]
\centering
\includegraphics[width=\textwidth]{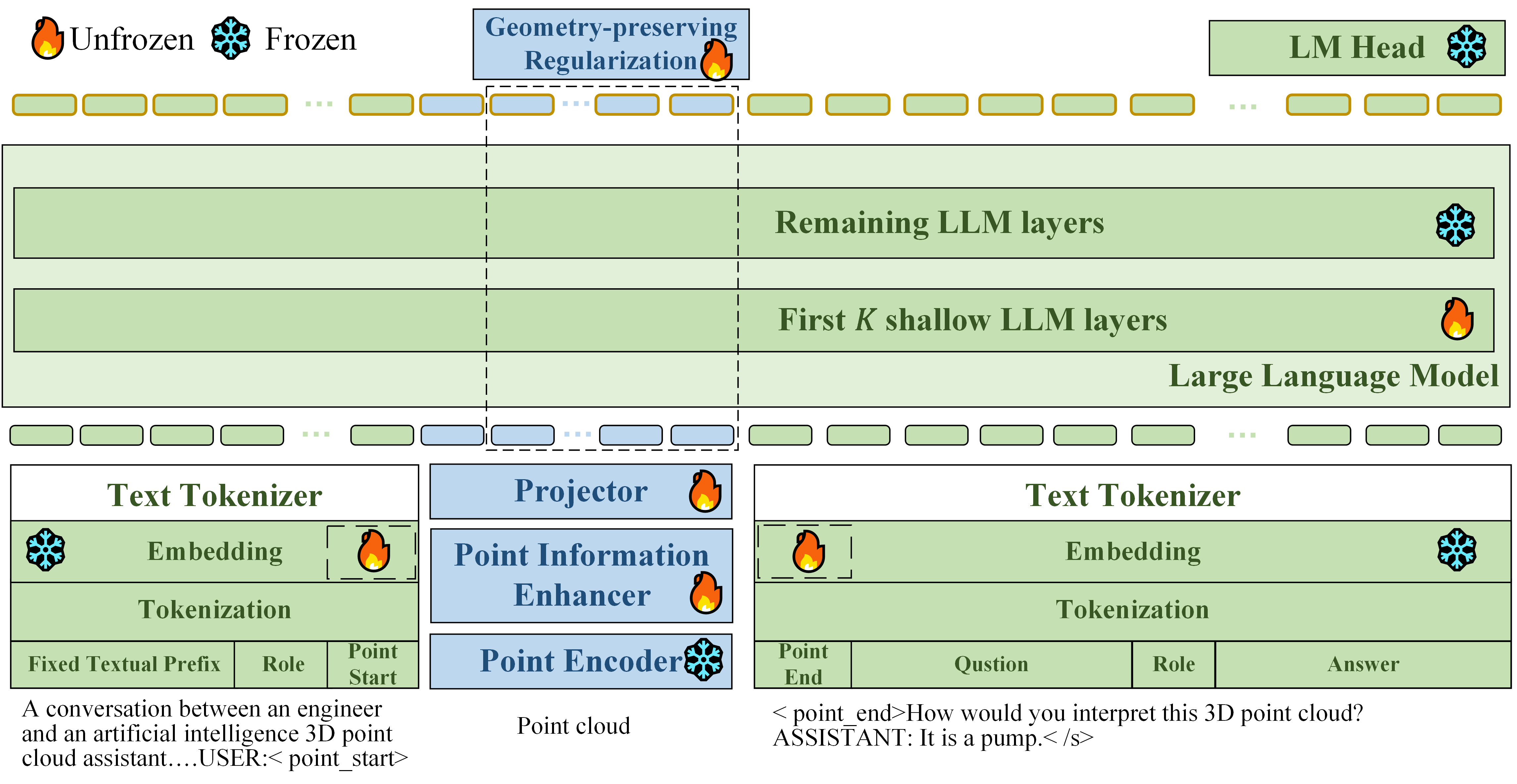}
\caption{Building-MLLM architecture for geometry-preserving cross-modal alignment in Stage 1.}
\label{F4B}
\end{figure}

\autoref{F4B} illustrates the Building-MLLM framework with geometry-language alignment training strategy. We use point clouds and their Simple Recognition text instruction-following  to establish alignment between the point cloud and language embedding spaces. To ensure stable modality representations, the pretrained Point Encoder is frozen to preserve low-level geometric features, and the Tokenizer embedding layer is frozen to maintain a consistent linguistic space. The Projector is trained to map point cloud features into the language space, and the two special tokens \texttt{<point\_start>} and \texttt{<point\_end>} are trained to help the LLM identify the point cloud segment within the mixed sequence. The above remains consistent with PointLLM, while our method differs in that we introduce a trainable PIE to provide task-oriented geometric and semantic enhancement before projection. In addition, we unfreeze the first K shallow LLM layers and apply GPR exclusively to these to prevent fine-grained geometric details from being oversmoothed during geometry-language alignment.

\subsubsection{Text Tokenizer}
The Tokenizer in Building-MLLM converts domain-specific input text into token IDs, which are then mapped into continuous representations compatible with the LLM. As shown in \autoref{F4B}, text input consists of the following components: \texttt{"Fixed Textual Prefix"}, \texttt{"Role"}, \texttt{"<point\_start>"}, \texttt{"<point\_end>"}, \texttt{"Question"}, \texttt{"Role"}, and \texttt{"Answer"}. 
The \texttt{"Fixed Textual Prefix"} is a handcrafted prompt designed to guide the model toward domain-specific behavior and to stabilize stylistic consistency during generation:

\begin{quote}\small
\texttt{A chat between a curious user and a civil engineering artificial intelligence assistant. The assistant gives helpful, detailed, and polite answers.}
\end{quote}

The \texttt{"Role"} denotes either \texttt{"USER"} or \texttt{"ASSISTANT"}. The \texttt{"<point\_start>"} and \texttt{"<point\_end>"} tokens serve as explicit boundaries for the point-cloud segment. They replace the single placeholder \texttt{"point"} (in \autoref{T4}) after flattening mechanism. \texttt{"Question"} and \texttt{"Answer"} correspond to the textual instruction and its expected response, respectively.

The Tokenizer first converts the input text sequence into vocabulary indices, 
where regular tokens fall within the range \([0, 31999]\), and the special 
\texttt{"<point\_start>"} and \texttt{"<point\_end>"} are assigned indices 32001 and 32002, respectively. 
These integer indices are then embedded into continuous representations compatible with the LLM:

\begin{equation}
E_s = \text{Tokenizer}(T) \in \mathbb{R}^{B \times L \times D}
\end{equation}

where \(T\) denotes the textual input to the Tokenizer, \(B\) is the batch size, 
\(L\) is the maximum sequence length within the batch, and \(D\) is the feature dimension, shared with the LLM hidden size (e.g., 4096).

\subsubsection{Point Information Enhancer (PIE)}

\begin{figure}[!htb]
\centering
\includegraphics[width=\textwidth]{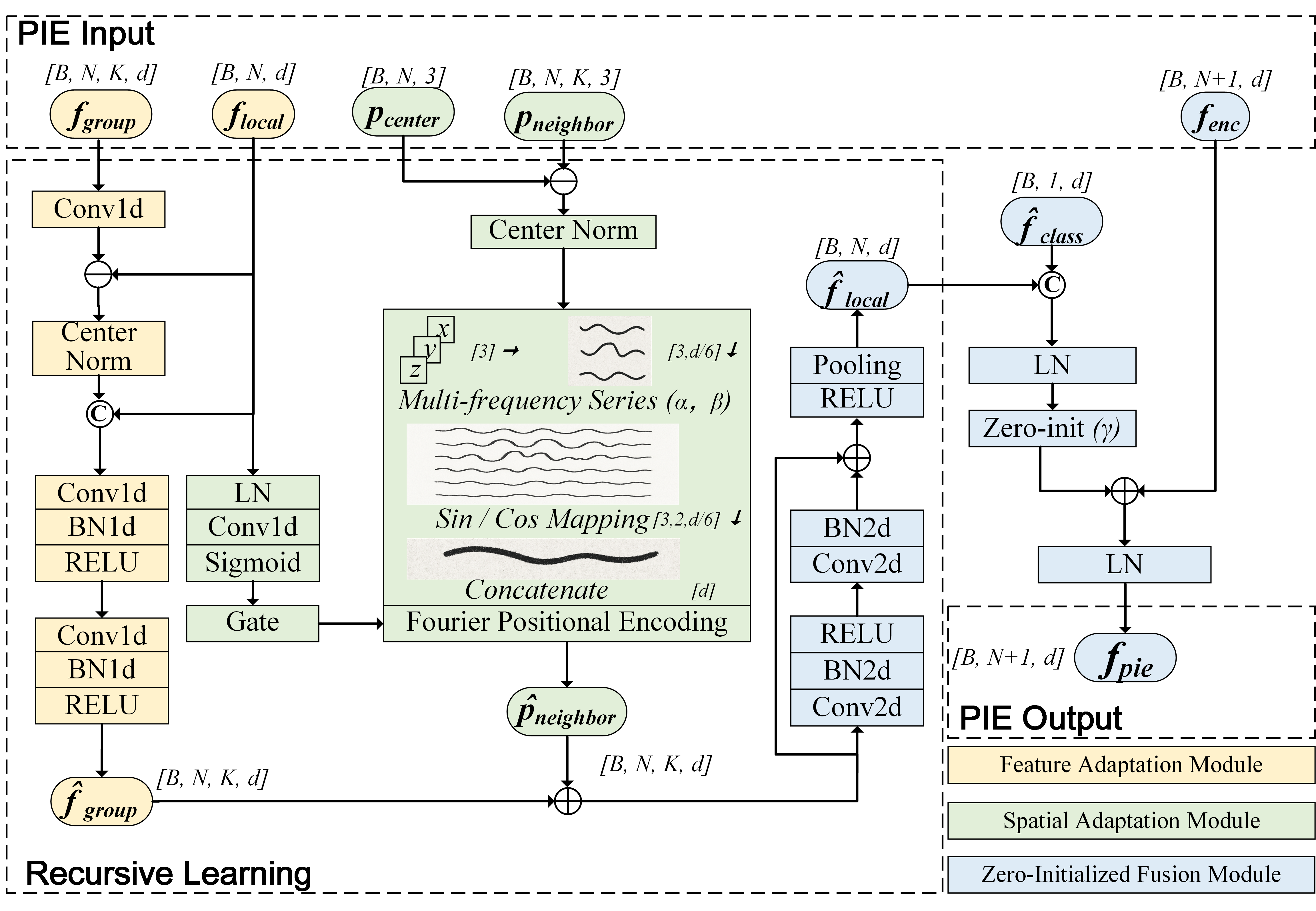}
\caption{Point Information Enhancer (PIE) module for task-oriented geometric enhancement.}
\label{F5}
\end{figure}

 The frozen point encoder, inherited from Point-BERT and consistent with the baseline PointLLM, is pretrained on large-scale datasets, models point clouds as token sequences, and provides stable, generalizable, and text-aligned geometric primitives. PIE is designed to retain these low-level geometric cues while enhancing task-relevant local discrimination and spatial contextualization. As shown in \autoref{F5}, given the encoder output 
\(f_{\text{enc}} \in \mathbb{R}^{B \times (N+1) \times d}\) containing 
\(N\) point tokens and one global token with \(d\)-dimensional 
features, PIE further consumes four 
complementary inputs. The local point features 
\(f_{\text{local}} \in \mathbb{R}^{B \times N \times d}\), obtained by 
removing the global token, provide point-wise semantic cues, while the 
grouped neighborhood features 
\(f_{\text{group}} \in \mathbb{R}^{B \times N \times K \times d}\) capture 
multi-point geometric context. In addition, the center coordinates 
\(p_{\text{center}}\) and neighborhood coordinates 
\(p_{\text{neighbor}}\) deliver explicit 3D spatial information for 
positional modeling.

\textbf{Feature Adaptation Module} aims to enrich point-level contextual 
dependencies and geometric relationships, thereby enhancing the model's ability to discriminate subtle 
local variations among building components. Specifically, \(f_{\text{group}}\) is first passed through a linear projection, while 
\(f_{\text{local}}\) is aligned through broadcasting. The two are then 
subtracted and center-normalized. The resulting 
features are concatenated with \(f_{\text{local}}\) and processed by a two-stage 1D convolutional block 
(Conv–BN–ReLU), with the first stage performing channel reduction 
and the second restoring the original dimensionality.We abstract this process as a function \(H(\cdot)\):

\begin{equation}
\hat{f}_{\text{group}} = H(f_{\text{local}},\, f_{\text{group}}),
\quad 
\hat{f}_{\text{group}} \in \mathbb{R}^{B \times N \times K \times d}.
\end{equation}

\textbf{Spatial Adaptation Module} enhances geometric awareness through
a semantically gated Fourier positional encoding. \(p_{\text{center}}\) and
\(p_{\text{neighbor}}\) are first subtracted and normalized and mapped into a
multi-frequency space using exponential bases controlled by parameters
$\alpha$ and $\beta$, producing frequency components that span both
coarse and fine spatial scales. These components are then transformed by
sine and cosine functions and concatenated to form a high-dimensional
positional representation with improved sensitivity to local geometric
patterns. To introduce semantic adaptivity, 
$f_{\text{local}}$ are used to generate a gating weight
$G(f_{\text{local}})\in[0,1]$, which modulates the Fourier features in a
channel-wise manner. This gating mechanism dynamically adjusts the
strength of positional encoding across different regions, enabling the
model to emphasize informative spatial cues while suppressing less
relevant ones:
\begin{equation}
\hat{p}_{\text{neighbor}} =
G(f_{\text{local}})\cdot
\mathrm{Fourier}(p_{\text{center}},\, p_{\text{neighbor}}),
\quad
\hat{p}_{\text{neighbor}} \in \mathbb{R}^{B\times N\times K\times d}.
\end{equation}

\textbf{Zero-Initialized Fusion Module} integrates the newly learned 
features into the encoder representation in a gradual and stable manner. After fusing $\hat{f}_{\text{group}}$ and $\hat{p}_{\text{neighbor}}$, which are the outputs of the Feature Adaptation Module and the Spatial Adaptation Module, respectively, the combined features are processed by a residual bottleneck block that compresses, filters, and strengthens the fused information, followed by a pooling operation to obtain a stable point-wise representation 
$\hat{f}_{\text{local}}$. This procedure forms a recursive learning mechanism, which consists of the Feature Adaptation Module, the Spatial Adaptation Module, and the partial Zero-Initialized Fusion Module up to the generation of $\hat{f}_{\text{local}}$. The number of recursive blocks within this mechanism is a configurable design choice. Across iterations, the resulting $\hat{f}_{\text{local}}$ is fed back into the original $f_{\text{local}}$, progressively refining the semantic representations.  $\hat{f}_{\text{local}}$ is then concatenated with a learnable global token $\hat{f}_{\text{class}}$ to match the dimensionality of the encoder 
output $f_{\text{enc}}$, and normalized by LayerNorm. To ensure a stable interaction with the backbone encoder, we introduce a single, global learnable scaling parameter $\gamma$, which is initialized to zero and applied only after the entire recursive learning process. This design guarantees that the recursive learning branch responsible for newly learned features introduces no perturbation to the frozen encoder representation at the early stage of training. As training progresses, the contribution of the enhanced features is gradually increased as $\gamma$ is updated, allowing the recursively refined representations to be smoothly injected into the backbone. The scaled enhanced features are then added to $f_{\text{enc}}$ and normalized again, producing the final PIE representation:

\begin{equation}
f_{\text{pie}} = \mathrm{LN}\!\left(
f_{\text{enc}} + \gamma \cdot \mathrm{LN}\!\left(
\left[ \hat{f}_{\text{class}} \,\|\, \hat{f}_{\text{local}} \right]
\right)
\right),
\quad f_{\text{pie}} \in \mathbb{R}^{B \times (N+1) \times d}
\end{equation}

\subsubsection{Geometry-Preserving Regularization (GPR)}

The Projector maps the $d$-dimensional ($384$) $f_{\text{pie}}$ 
 produced by PIE into the LLM textual embedding space of 
dimension $D$ ($4096$):
\begin{equation}
E_{\text{pt}} = \mathrm{Projector}(f_{\text{pie}})
\in \mathbb{R}^{B \times (N+1) \times D}.
\end{equation}

\begin{figure}[!htb]
\centering
\includegraphics[width=0.8\textwidth]{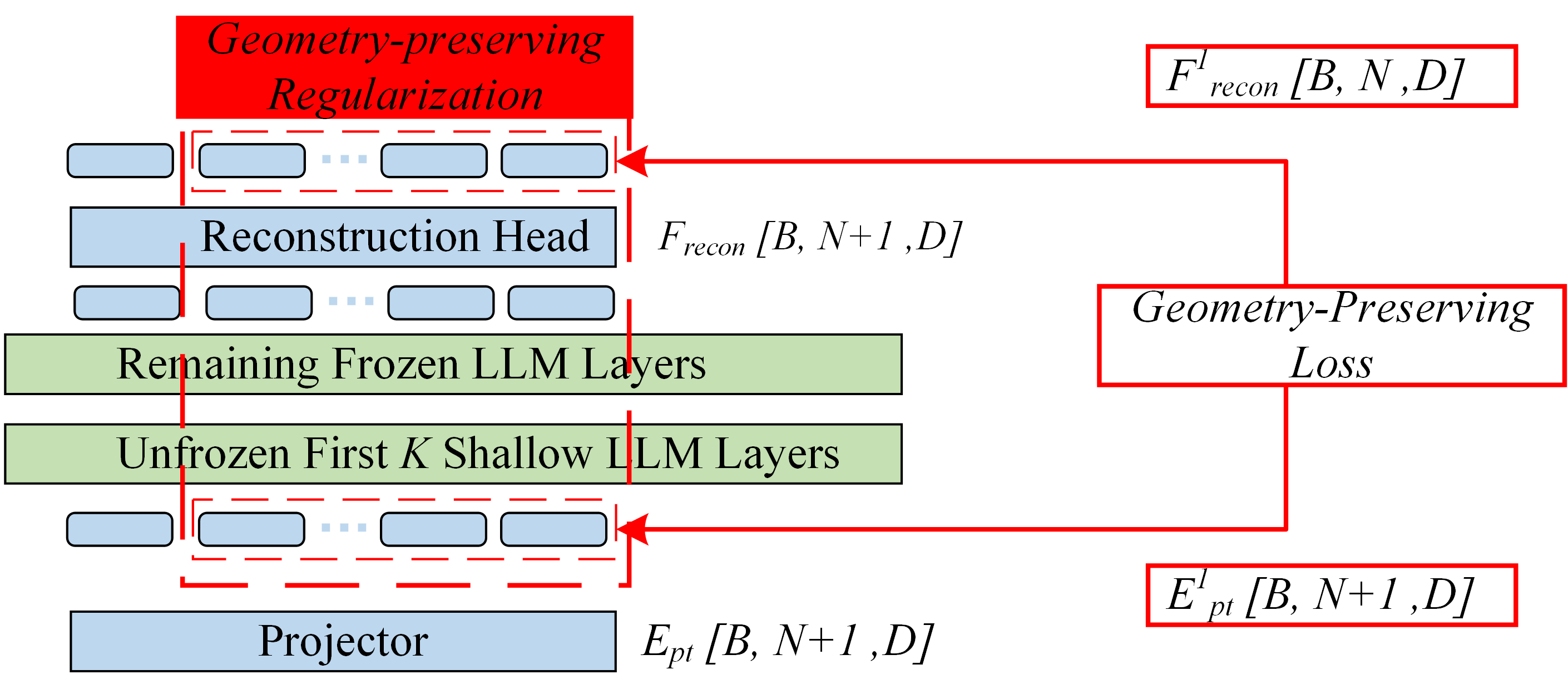}
\caption{Proposed Geometry-Preserving Regularization (GPR) on shallow LLM layers.}
\label{F5B}
\end{figure}

Indoor building components often exhibit subtle geometric differences—a 90$^{\circ}$ elbow versus a 45$^{\circ}$ elbow, or a gate valve versus a globe valve—that carry critical semantic meaning but are difficult to preserve during cross-modal alignment. Geometric erosion describes the tendency for detailed 3D geometric representations in point clouds to be progressively absorbed into higher-level semantic abstractions, thereby reducing the model's sensitivity to true geometric details. The projected point cloud features $E_{\text{pt}}$ encode these fine-grained geometric patterns. However, when these features are concatenated with textual embeddings and processed by the LLM, the semantic dominance of the language modality tends to pull $E_{\text{pt}}$ toward the higher-level abstraction space $E_{\text{s}}$. This alignment pressure may oversmooth low-level geometric details, thereby weakening discrimination performance on visually similar components. To counteract this effect, we introduce Geometry-Preserving Regularization (GPR) illustrated in \autoref{F5B}, which enforces geometric fidelity by applying a reconstruction-based constraint exclusively to the shallow unfrozen LLM layers, ensuring that essential structural cues remain accessible throughout the model pipeline.

$E_{\text{pt}}$ participates in the mixed 
sequence and is processed by the LLM, after which a Reconstruction Head 
produces the reconstructed features $F_{\text{recon}}$ with the same 
dimensionality. To impose local geometric constraints on shallow unfrozen 
layers, we remove the global token and use only the patch tokens: 
layers, we remove the global token and use only the patch tokens: 
$E_{\text{pt}}^{1}\in \mathbb{R}^{B \times N \times D}$ is treated as the ground truth and 
$F_{\text{recon}}^{1}\in \mathbb{R}^{B \times N \times D}$ as the prediction.

The geometry-preserving loss is 
defined as the mean squared error (MSE) between the two across all 
batches and patches:

\begin{equation}
\mathcal{L}_{\text{GPR}} 
= \frac{1}{BN} 
\sum_{b=1}^{B} \sum_{n=1}^{N} 
\frac{1}{D} \sum_{d=1}^{D}
\left( F_{\text{recon}}^{1} - E_{\text{pt}}^{1} \right)^{2}.
\end{equation}

The overall training objective of this stage is defined as follows:

\begin{equation}
\mathcal{L}_{\text{stage1}}
= \mathcal{L}_{\text{LM}} + \mathcal{L}_{\text{GPR}}
= - \sum_{t \in \mathcal{T}_{\text{resp}}} \log P(y_t \mid y_{<t}, Y) + \mathcal{L}_{\text{GPR}}.
\end{equation}

where $\mathcal{T}_{\text{resp}}$ denotes the set of token positions
corresponding to the model's responses. $\mathcal{L}_{\text{LM}}$ is computed in an
autoregressive manner, encouraging the LLM to predict each target token
$y_t$ given all preceding tokens $y_{<t}$ and the multimodal input $Y$,
while masking out the user-provided instruction tokens.For $\mathcal{L}_{\text{GPR}}$, its gradients 
are detached before reaching the Projector, 
ensuring that the geometric constraint is applied only to the unfrozen shallow 
LLM layers.

\subsection{Stage 2: Multi-Task Instruction Fine-Tuning}

\begin{figure}[!htb]
\centering
\includegraphics[width=\textwidth]{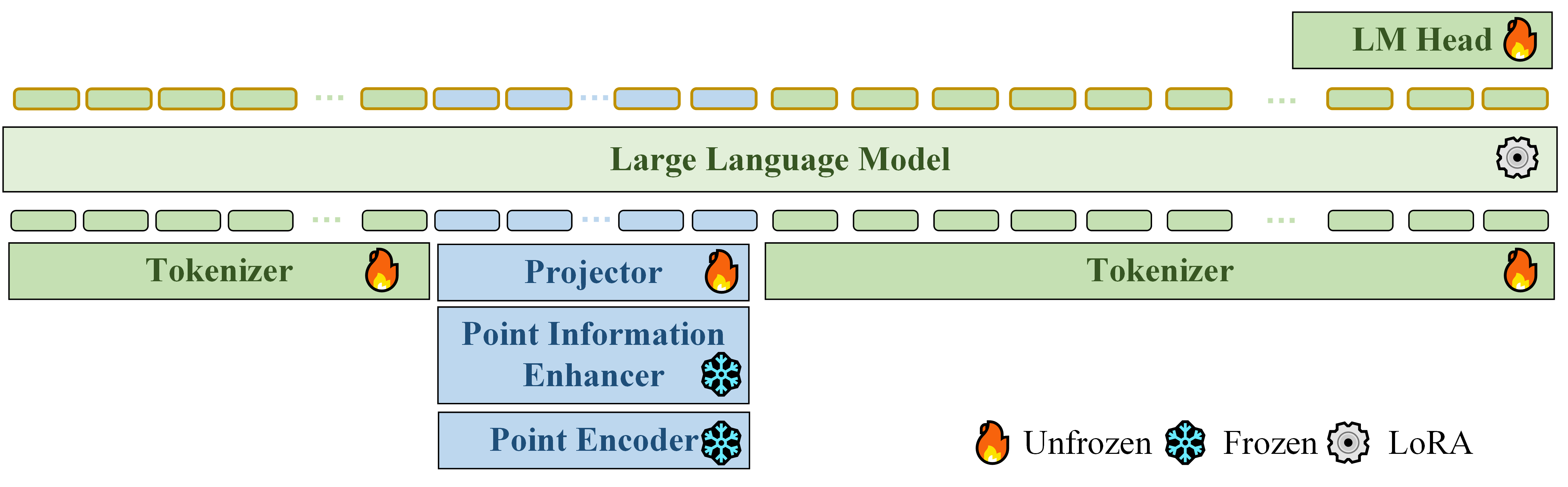}
\caption{Proposed Building-MLLM framework with stage-2 instruction fine-tuning training.}
\label{F4C}
\end{figure}

In contrast to stage 1, which focuses on aligning point cloud with textual semantics, stage 2 aims to equip the model with cross-modal generation capability, enabling natural-language recognition, description, and reasoning over geometric inputs. This stage emphasizes generative understanding that transforms geometric structures into semantic expressions under instruction guidance. During this stage, we jointly train the model on indoor component point clouds and their corresponding instruction-following from the Simple Recognition, Complex Captioning, and Multi-Engineering QA datasets. The Stage-2 training loss $\mathcal{L}_{\text{stage2}}$ is essentially the $\mathcal{L}_{\text{LM}}$. While PointLLM fully unfreezes all modules except the Point Encoder, Building-MLLM introduces two modifications, as shown in \autoref{F4C}. First, the newly added PIE is kept frozen to maintain the geometric-linguistic alignment established in stage 1. Second, we apply a multi-dimensional optimized LoRA to the LLM backbone.This preserves the model's inherent competence while providing low-rank adaptability for cross-modal generation, enabling efficient domain adaptation to indoor components, balanced multi-task learning, and reduced training cost.

\begin{figure}[!htb]
\centering
\includegraphics[width=\textwidth]{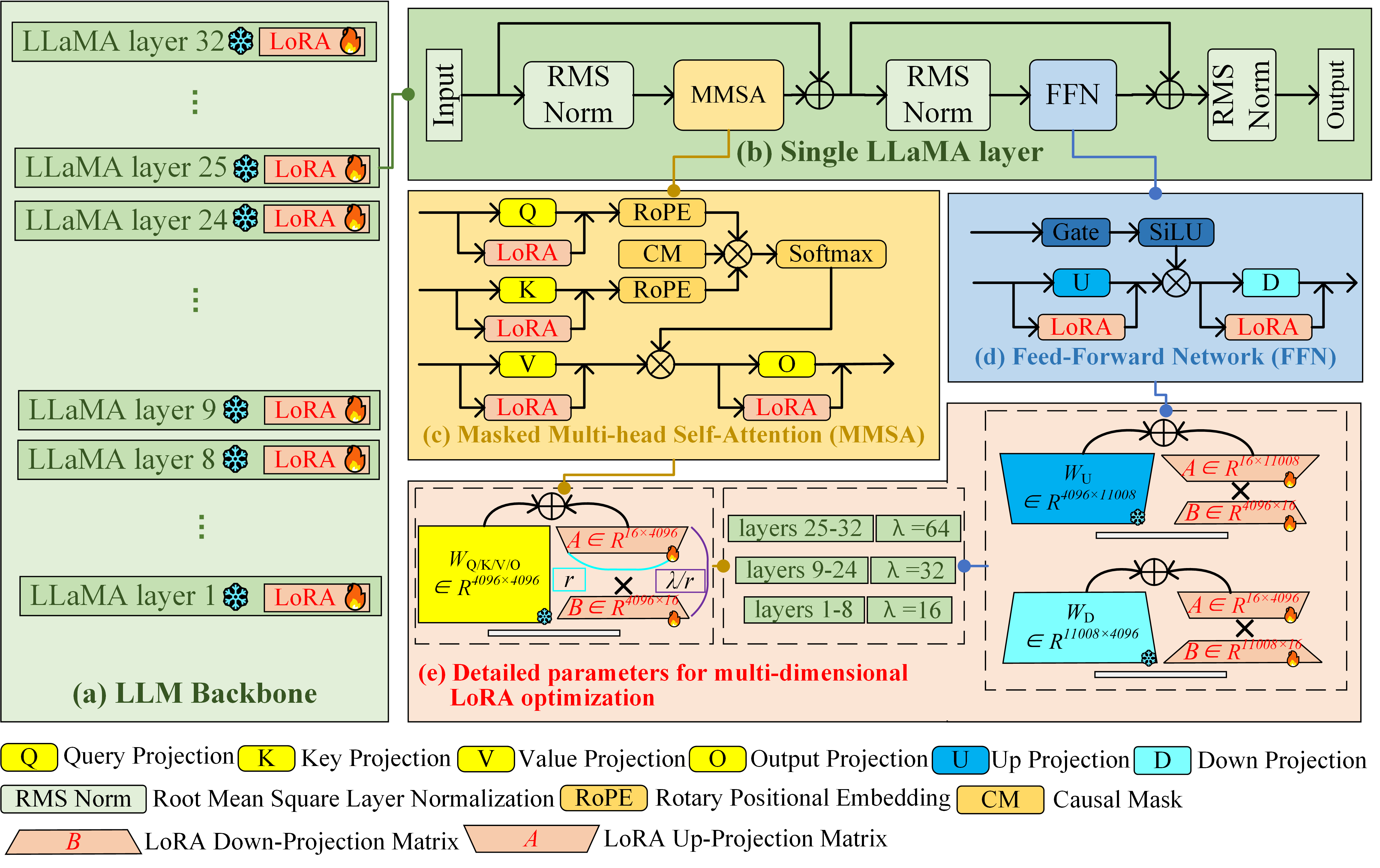}
\caption{Multi-Dimensional LoRA fine-tuning Strategy for Building-MLLM.}
\label{F6}
\end{figure}

 Building-MLLM adopts LLaMA-7B as its LLM backbone, comprising 32 stacked Transformer layers. As a decoder-only architecture, LLaMA demonstrates strong semantic modeling, contextual coherence, and long-range dependency capture, providing a stable foundation for LoRA insertion and cross-modal adaptation as shown in \autoref{F6}(a). Each LLaMA layer, illustrated in \autoref{F6}(b), contains a Masked Multi-Head Self-Attention (MMSA) module for modeling intra-sequence dependencies and a Feed-Forward Network (FFN) for nonlinear transformation and semantic enhancement—together with Root Mean Square Layer Normalization (RMSNorm) to stabilize deep training.

For fine-tuning, we apply LoRA to the linear transformations of LLaMA. Its formulation is:

\begin{equation}
W' = W + \frac{\lambda}{r} \cdot \Delta W
   = W + \frac{\lambda}{r} \cdot BA,
\qquad 
B \in \mathbb{R}^{d_{\text{in}} \times r},\;
A \in \mathbb{R}^{r \times d_{\text{out}}},
\qquad r \ll d_{\text{in}}, d_{\text{out}}.
\end{equation}

Here, $W$ denotes the original weight matrix, while $B$ and $A$ constitute a rank-$r$ residual update, and $\lambda$ is a scaling factor. By freezing $W$ and training only the low-rank residual branch $BA$, LoRA markedly reduces the number of trainable parameters and enables efficient adaptation to new tasks and cross-modal distributions, without compromising the model's inherent language capabilities. However, the effectiveness of LoRA within an LLM backbone that tightly couples geometric and linguistic representations is not guaranteed a priori. To fully leverage its potential in multimodal settings—balancing fine-grained geometric recognition with long-range semantic reasoning, and supporting heterogeneous tasks such as Simple Recognition, Complex Captioning, and Multi-Engineering QA—we systematically examine key LoRA configuration dimensions:

\textbf{ Layer Range.} Determines the depth-wise placement of LoRA modules, influencing geometric sensitivity, semantic abstraction, and the hierarchy of cross-modal fusion.

\textbf{ Target Modules.} Applying LoRA to MMSA projection and FFN projection jointly shapes cross-modal capacity, semantic recomposition flexibility, and contextual dependency modeling strength.

\textbf{ Rank $r$.} Controls the capacity of the low-rank update, affecting expressive power, stability, and performance on long-text and reasoning-intensive tasks.

\textbf{ Scaling Strategy $\lambda$.} Regulates the magnitude of LoRA updates across layers, affecting task balance and the convergence dynamics of geometry–semantic adaptation.

our final configuration adopts full 32-layer coverage \autoref{F6}(a), joint injection into q/k/v/o and FFN u/d \autoref{F6}(c,d), a rank of 16, and a progressively increasing scaling schedule (front 8 layers: 16, middle 16 layers: 32, final 8 layers: 64) as shown in \autoref{F6}(e), yielding balanced improvements in geometric discrimination, semantic organization, and long-range reasoning for indoor building components. The experimental process is presented in Section \ref{experiments}.

\section{Experiments}
\label{experiments}

This section presents a comprehensive experimental evaluation of Building-MLLM on our point cloud-text instruction-following benchmark. We report both qualitative and quantitative results to assess the model's performance across Simple Recognition, Complex Captioning, and Multi-Engineering QA tasks. Specifically, we first describe the dataset split, implementation details, and training protocol, then introduce the evaluation framework, followed by qualitative studies and systematic quantitative ablations. In addition, we present case-based analyses of the evaluation framework and inference case studies conducted on real-world point cloud datasets.

\subsection{Experimental Setup}

In this study, we use our point cloud-text instruction-following dataset, which contains 4,198 point cloud objects and 37,782 instruction-following data. Among them, 800 point clouds and 7,200 text instruction-following pairs that were manually reviewed and corrected are used for evaluation, and are evenly divided into two subsets of equal size (400 objects each). The first subset is used as the validation inference set for baseline model screening and hyperparameter tuning in indoor building scenes, while the second subset serves as the test inference set for the final, independent performance evaluation. The remaining 3,398 point clouds and their corresponding 30,582 text instruction-following pairs constitute the training set, including 3,398 Simple Recognition, 3,398 Complex Captioning , and 23,786 multi-engineering QA data. This split ensures strict isolation between training, validation, and testing data while maintaining distributional consistency across different phases.

All experiments are conducted on 2\(\times\)NVIDIA RTX A6000 GPUs (52 GB per card) with Driver~535.230.02 and CUDA~12.2, using PyTorch 2.7.1 and the Transformers 4.40.2 framework. To reduce memory consumption and enhance throughput, we enable FlashAttention, bfloat16 (bf16) mixed precision, and gradient checkpointing.

Training follows a two-stage strategy. In Stage~1, geometry-language alignment is performed using the point cloud and  Simple Recognition instructions from the training set for 3~epochs with a per-GPU batch size of 8. The initial learning rate is set to \(2\times10^{-3}\), optimized with AdamW ( weight decay~=~0.05), cosine annealing learning rate scheduling, and a warmup ratio of 0.03. The maximum sequence length is set to 2048. The resulting weights from Stage~1 are used to initialize Stage~2. In Stage~2, instruction finetuning is performed using all three task types for 2~epochs with a per-GPU batch size of~4 and an initial learning rate of \(3\times10^{-5}\), again using AdamW with cosine annealing scheduling. The training process remains stable across both stages and consistently converges under the multi-task setting.

Regarding model scale and computational requirements, we emphasize that the LLaMA-7B backbone is adopted as a research-stage reference model to validate the feasibility and upper-bound capability of the proposed framework, rather than as a fixed configuration for practical deployment. Notably, the 7B backbone parameters are never fully trained in either stage. In practice, the peak GPU memory consumption during training is approximately 38 GB per card, while inference requires around 30 GB, which is feasible in typical engineering research environments and significantly lower than full backbone training paradigms. For potential real-world deployment, point cloud encoding can be performed locally or at the edge, while language-based reasoning and response generation can be executed in an offline or asynchronous manner on centralized servers. This decoupled deployment strategy aligns well with common engineering system architectures and supports practical integration into intelligent facility management and digital twin workflows.

System performance is evaluated using a combination of traditional automatic metrics and GPT-4–based high-level semantic evaluation metrics. The detailed evaluation metrics are provided in the following subsection.

\subsection{Evaluation Framework and Metrics}

Given the openness and expressive diversity of MLLM outputs, we adopt two categories of automatic metrics: lexical overlap metrics (BLEU, ROUGE, METEOR) to measure word-level similarity, and semantic similarity metrics (Sentence-BERT, SimCSE) to assess sentence-level semantic proximity. These metrics are commonly used for captioning tasks but remain limited in capturing diverse yet semantically equivalent expressions as well as factual consistency or domain-specific engineering logic. To address these limitations, we introduce GPT-4 as a high-level semantic evaluator, whose judgments have been shown to closely align with human experts and to handle semantic equivalence beyond surface-level variations capabilities that traditional metrics lack. In addition to retaining automatic metrics for the Complex Captioning task, we design task-specific standardized GPT-4 evaluation prompts tailored to different task types. These prompts explicitly target the core objectives of each task, including recognition correctness, attribute completeness, engineering logic consistency, and factual accuracy, enabling a more comprehensive and reliable assessment across heterogeneous tasks.

It should also be noted that the proposed Building-MLLM, GPT-4V used for data generation, and GPT-4 employed in the evaluation stage played different roles in the pipeline, with distinct input conditions and task objectives. In addition, we introduced multiple constraint strategies throughout the study to minimize potential bias and to preserve as much data independence as possible. Nevertheless, we conservatively acknowledge that some bias may still remain, and systematically quantifying this potential bias will be an important direction for future work.

\newpage

For the Simple Recognition task, we use a unified input instruction ("What is it?") to obtain the model's natural-language prediction. GPT-4 then evaluates whether the model output refers to the same object type as the verified label, irrespective of surface-level wording differences. The GPT-4 evaluation prompts designed for this task are illustrated in \autoref{T6}, GPT-4 provides a binary "T/F'' decision along with a concise rationale. This setup highlights GPT-4's ability to capture type-level semantic equivalence while ignoring superficial phrasing differences, ensuring consistent and reproducible evaluation across samples.

\begin{table}[!htb]
\footnotesize
\captionsetup{margin=0cm}
\caption{Designed standardized GPT-4 evaluation prompt for the Simple Recognition task.}
\centering
\begin{tblr}{
  hline{1-6} = {-}{},
  colspec={Q[1.5cm] Q[14.5cm]},
  cell{2}{2} = {valign=t},
}
\textbf{Prompt}  & {Analyze two sentences and determine if they're referring to the same general object or concept, focusing on the type of object only, not attributes. Respond with "T" if they refer to the same thing and "F" if not. Also, provide a brief rationale (no more than 30 words) for your judgment.\\ \newline \textbf{Example:} \\Input: 1.A lateral pipe junction fitting. 2.~You are looking at a lateral pipe junction fitting.\\Output:T\#Both refer to the same type of pipe fitting.\newline \\Now, analyze the following:\\Input: 1. \{ground\_truth\}  2. \{model\_output\}\\Output:~}                                                                                                                                                                                                                                  \\
\textbf{Example} & {Input: 1.This is a lateral pipe junction fitting point cloud. 2.~This is a pipe bend fitting point cloud.\\Output:F\#Although both sentences refer to a fitting point cloud, the first refers to a lateral pipe junction while the second refers to a pipe bend.}                                                                                                                                                                                                                     \\ 
\end{tblr}
\label{T6}
\end{table}

\newpage

For the Complex Captioning task, we use an input instruction ("Caption this 3D model in detail.") to obtain the model prediction and design task-specific GPT-4 evaluation prompts shown in \autoref{T7}.  GPT-4 first parses the human caption into key semantic aspects—category, geometry, structural features, functional usage, connection form, and distinguishing attributes—and then checks whether these aspects are fully or partially expressed in the model-generated caption. GPT-4's strong semantic understanding and reasoning ability enable it to recognize diverse yet semantically equivalent expressions, cross-sentence correspondences, and logical consistency. Each aspect contributes equally to the final 0–100 score, ensuring a standardized and interpretable caption-level evaluation that overcomes the limitations of traditional metrics.

\begin{table}[!htb]
\footnotesize
\captionsetup{margin=0cm}
\centering
\caption{Designed standardized GPT-4 evaluation prompt for the Complex Captioning task.}
\renewcommand{\arraystretch}{1.2}
\begin{tabularx}{\textwidth}{p{0.1\textwidth}X}
\hline
\textbf{Prompt} & 
Evaluate a model-generated caption against a human-generated caption (ground truth) for a 3D model. First, extract the key aspects mentioned in the human caption. Possible aspects include, but are not limited to: category, shape, structural features, function/usage, installation/connection, distinguishing attributes. \newline

Be generous when judging similarity. Score from 0 to 100, where each aspect contributes equally to the score. Consider similar concepts for partial score.
Provide your score (0-100) and a short justification (less than 30 words) in the format of "score\#reason". \newline

\textbf{Example:} \newline
Human: The object is a railing with vertical balusters and horizontal top and bottom rails. It features a linear, elongated structure with evenly spaced vertical elements. The connection forms include welded joints, providing stability and rigidity. Its distinguishing feature is the uniform spacing of balusters, ensuring safety and aesthetic appeal. \newline
Model: The object is a railing with a sturdy rectangular top and side panels. It has a linear geometry and evenly spaced vertical bars, creating a balanced structure. The railing offers safety and support, typically found in staircases or boundaries. \newline
Output: 90\#mentioned railing category, structural features, function, distinguishing attributes; geometry similar; connection missing. \newline

Now score the following: \newline
Human: \{ground\_truth\} \newline
Model: \{model\_output\} \newline
Output: \\ \hline

\textbf{Example} & 
Human: The object is a duct bend fitting with a curved, angular structure. It features a smooth, semi-circular geometry with flanged ends for connection. The fitting is designed to redirect airflow efficiently within duct systems, characterized by its robust, metallic appearance and precise angles. \newline
Model: The object is a duct bend fitting, featuring a rectangular shape with a 90-degree bend. It has a smooth surface and angular geometry. This fitting connects ductwork sections, facilitating a change in direction. The bend is gentle and non-hazardous, ideal for home HVAC systems. \newline
Output: 70\#mentioned duct bend fitting category, shape, structural features, function/usage; geometry similar; connection mentioned with related concept of connecting ductwork sections. \\ \hline

\end{tabularx}
\label{T7}
\end{table}

\newpage

For the Multi-engineering QA task, we employ seven engineering types of questions, each with kinds of distinct instruction form dataset guiding the model to generate task-specific textual answers. Using the task-specific evaluation prompts in \autoref{T8}, GPT-4 evaluates the model answer by aligning it with the key engineering points present in the verified response, taking into account the task context rather than simple answer coverage. This allows GPT-4 to assess semantic consistency, engineering logic correctness, and adherence to task intent—providing a more reliable and targeted evaluation for professional engineering QA scenarios.

\begin{table}[!htb]
\centering
\footnotesize
\captionsetup{margin=0cm}
\caption{Designed standardized GPT-4 evaluation prompt for the Multi-engineering QA task.}
\renewcommand{\arraystretch}{1.2}
\begin{tabularx}{\textwidth}{p{0.1\textwidth}X}
\hline
\textbf{Prompt} & 
Here is \{task\_subtype\} task.Evaluate a model-generated answer against a human-generated answer (ground truth) for a 3D model.Identify the aspects mentioned in the human answer, and calculate the percentage of correctly mentioned or matched in the model answer. Please give a score from 0 to 100 based on how well the model answer matches the ground truth in the format of 'score\#reason'. \newline

\textbf{Example:} \newline
Question: What is the function of this object?\newline
Human: To change airflow direction in HVAC systems. \newline
Model:  To redirect airflow in duct systems. \newline
Output: 100\#mentioned change airflow direction and semantically similar systems. \newline

Now score the following: \newline
Question: \{question\} \newline
Human: \{ground\_truth\} \newline
Model: \{model\_output\} \newline
Output: \\ \hline

\textbf{Example} & 
Question: Where should this kind object not be installed?\newline
Human: Avoid installing near obstructions that block airflow. \newline
Model: Installation in a spacious indoor pool or spa area. \newline
Output: 0\#The model answer does not address the aspect mentioned in the human answer about avoiding obstructions that block airflow. \\ \hline

\end{tabularx}
\label{T8}
\end{table}

\newpage

\subsection{Qualitative Analysis}

We first evaluate several SOTA MLLMs released by their original authors to compare their performance on indoor building components, focusing primarily on the Simple Recognition and Complex Captioning tasks. Following the input modality settings used in prior work, InstrucBLIP, LLaVA, and DreamLLM take single-view rendered images as input; 3D-LLM operates on multi-view rendered images together with point clouds; while MiniGPT-3D, ShapeLLM, and PointLLM directly use point clouds as their input modality.

\begin{table}[!htb]
\footnotesize
\captionsetup{margin=0cm}
\caption{Comparison of Simple Recognition performance against SOTA methods.}
\centering
\begin{tblr}{
  hline{1-2,9,11,13} = {-}{black},
}
No. & Model             & Input                   & Fine-tuning & GPT-4 \\
1   & InstrucBLIP       & Image       & \ding{55}   & 7.25   \\
2   & LLaVA             & Image       & \ding{55}   & 10.00  \\
3   & DreamLLM          & Image       & \ding{55}   & 9.25   \\
4   & 3D-LLM            & Point+Image & \ding{55}   & 12.25  \\
5   & MiniGPT-3D        & Point            & \ding{55}   & 23.50  \\
6   & ShapeLLM          & Point            & \ding{55}   & 14.50  \\
7   & PointLLM          & Point           & \ding{55}   & 28.25  \\
8   & PointLLM          & Point             & \ding{51}   & 82.00  \\
9   & Building-MLLM    & Point             & \ding{51}   & 88.75  \\
10  & \textbf{PointLLM*}         & Point            & \ding{51}   & 81.50  \\
11  & \textbf{Building-MLLM*}   & Point           & \ding{51}   & 88.00  \\
\end{tblr}
\label{T9}
\end{table}

Results on the validation inference set (see Experiments 1–7 in \autoref{T9} and \autoref{T10}) show that point cloud MLLMs consistently outperform image MLLMs across both tasks. This indicates that, in 3D understanding scenarios, point clouds—being the direct carrier of spatial geometry—provide stronger support for accurate recognition and semantic description. Moreover, due to the considerable domain gap between our indoor engineering scenario and existing open-domain datasets, all models exhibit lower absolute scores than those reported in their original papers, highlighting the inherent challenges brought by professional indoor engineering environments. Considering overall effectiveness and cross-scene robustness, we select PointLLM as the base model for further improvement and for initializing our two-stage training pipeline.

\begin{table}[!htb]
\footnotesize
\captionsetup{margin=0cm}
\caption{Comparison of Complex Captioning performance against SOTA methods.}
\centering
\begin{tblr}{
  hline{1-2,9,11,13} = {-}{},
}
No. & Model & Fine-tuning & Sentence-BERT & SimCSE & BLEU-1 & ROUGE-L & METEOR & GPT-4 \\
1   & InstrucBLIP       & \ding{55}   & 13.82 & 12.74 & 7.95 & 9.33 & 8.49 & 12.26 \\
2   & LLaVA             & \ding{55}   & 15.70 & 14.11 & 9.03 & 10.11 & 7.03 & 13.93 \\
3   & DreamLLM          & \ding{55}   & 15.19 & 14.62 & 9.91 & 10.41 & 8.11 & 13.48 \\
4   & 3D-LLM            & \ding{55}   & 17.42 & 16.16 & 8.73 & 11.29 & 8.99 & 15.29 \\
5   & MiniGPT-3D        & \ding{55}   & 23.58 & 27.36 & 15.93 & 16.68 & 15.74 & 22.03 \\
6   & ShapeLLM          & \ding{55}   & 19.24 & 15.87 & 8.00 & 12.19 & 6.12 & 17.10 \\
7   & PointLLM          & \ding{55}   & 29.05 & 25.41 & 18.41 & 17.62 & 17.71 & 24.36 \\
8   & PointLLM          & \ding{51}   & 62.12 & 66.96 & 24.36 & 24.63 & 18.02 & 59.27 \\
9   & Building-MLLM     & \ding{51}   & 65.76 & 70.48 & 29.06 & 31.71 & 24.98 & 64.92 \\
10  & \textbf{PointLLM*}         & \ding{51}   & 61.10 & 67.10 & 23.70 & 25.10 & 18.30 & 58.40 \\
11  & \textbf{Building-MLLM*}     & \ding{51}   & 65.30 & 70.70 & 28.70 & 32.00 & 25.10 & 65.10 \\
\end{tblr}
\label{T10}
\end{table}

Following this, Experiments 8–9 in \autoref{T9} and \autoref{T10} compare the performance of PointLLM and our proposed Building-MLLM after two-stage training on our dataset. In these experiments, PointLLM uses its default alignment mechanism in Stage 1 and adopts the original LoRA configuration in Stage 2. The results show that the architectural design and training strategy of Building-MLLM effectively enhance geometry-friendly geometry-language alignment and improve task robustness. Experiments 10–11 further evaluate the two models on the independent evaluation set, which does not participate in any parameter tuning and therefore better reflects true generalization ability. Our model achieves 88.00\% in Simple Recognition and 65.10\% in Complex Captioning, outperforming the baseline PointLLM by 7.50\% and 5.65\%, respectively. The performance gains on the independent evaluation set closely mirror the improvements observed on the validation set, indicating that the benefits of our method are transferable and not attributable to hyperparameter tuning or dataset-specific artifacts.

In \autoref{T11}, we compare the evaluation results of the Multi-engineering QA task. Using below test inference results as an example, our method demonstrates clear advantages. These improvements yield better performance in multi-turn engineering semantic understanding, constraint interpretation, and cross-sentence consistency. Specifically, the minimum accuracy among the seven sub-tasks increases from 35.20\% to 54.67\%, and the average accuracy improves from 59.57\% to 68.14\%. These gains indicate that the two-stage training strategy and the proposed LoRA design in Building-MLLM more effectively capture structured constraints and cross-turn semantic dependencies in engineering scenarios, resulting in stable and transferable improvements under complex QA settings.

\begin{table}[!htb]
\centering
\footnotesize
\captionsetup{margin=0cm}
\caption{Comparison of Multi-engineering QA performance against SOTA methods using GPT-4.}
\begin{tblr}{
  hline{1,2,4,6} = {-}{},
}
Model             & Input                   & Fine-tuning & Know & CoSe & Engi & FuCp & Cstr & SpRe & EmIn & Average \\
PointLLM          & Point            & \ding{51}   & 77.49  & 60.62  & 68.19  & 62.65  & 54.94  & 56.40  & 37.48  & 59.68   \\
Building-MLLM    & Point            & \ding{51}   & 80.93  & 71.35  & 73.68  & 68.86  & 66.87  & 69.95  & 53.30  & 69.28   \\
\textbf{PointLLM*}         & Point            & \ding{51}   & 78.32  & 60.71  & 66.98  & 67.45  & 54.50  & 53.89  & 35.20  & 59.57   \\
\textbf{Building-MLLM*}   & Point            & \ding{51}   & 79.65  & 71.47  & 71.24  & 69.26  & 66.01  & 64.74  & 54.67  & 68.14   \\
\end{tblr}

\begin{flushleft}
\footnotesize\textit{Abbreviations:}
\textit{Know} = Knowledge Capability;\ 
\textit{CoSe} = Commonsense Reasoning;\ 
\textit{Engi} = Advanced Engineering Reasoning;\ 
\textit{FuCp} = Functional Comparison;\ 
\textit{Cstr} = Constraint Reasoning;\ 
\textit{SpRe} = Spatial Relationship;\ 
\textit{EmIn} = Embodied Interaction.
\end{flushleft}

\label{T11}
\end{table}

\begin{figure}[!htb]
\centering
\includegraphics[width=\textwidth]{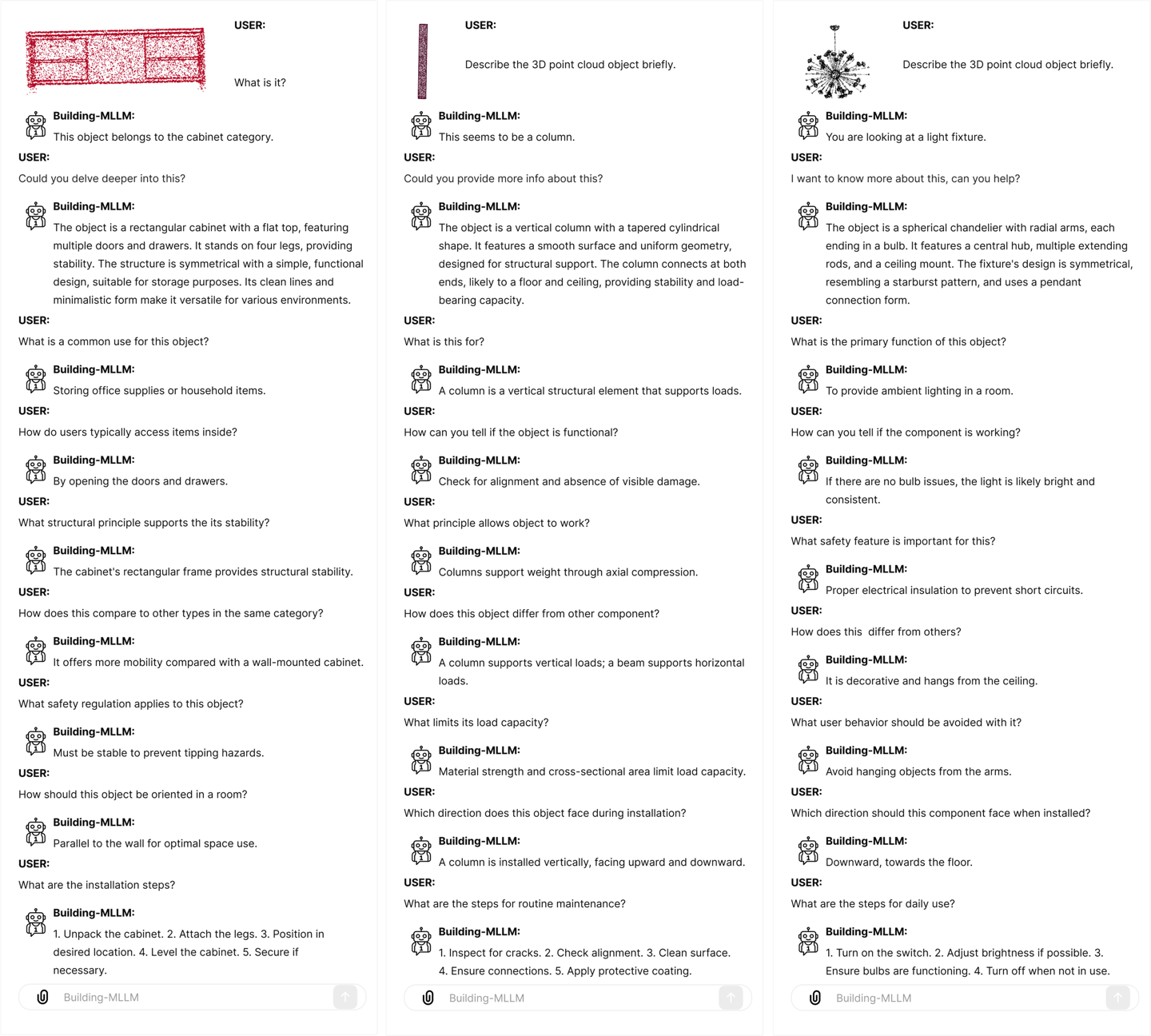}
\caption{Test case demonstration of Building-MLLM (1).}
\label{F8}
\end{figure}

\begin{figure}[!htb]
\centering
\includegraphics[width=\textwidth]{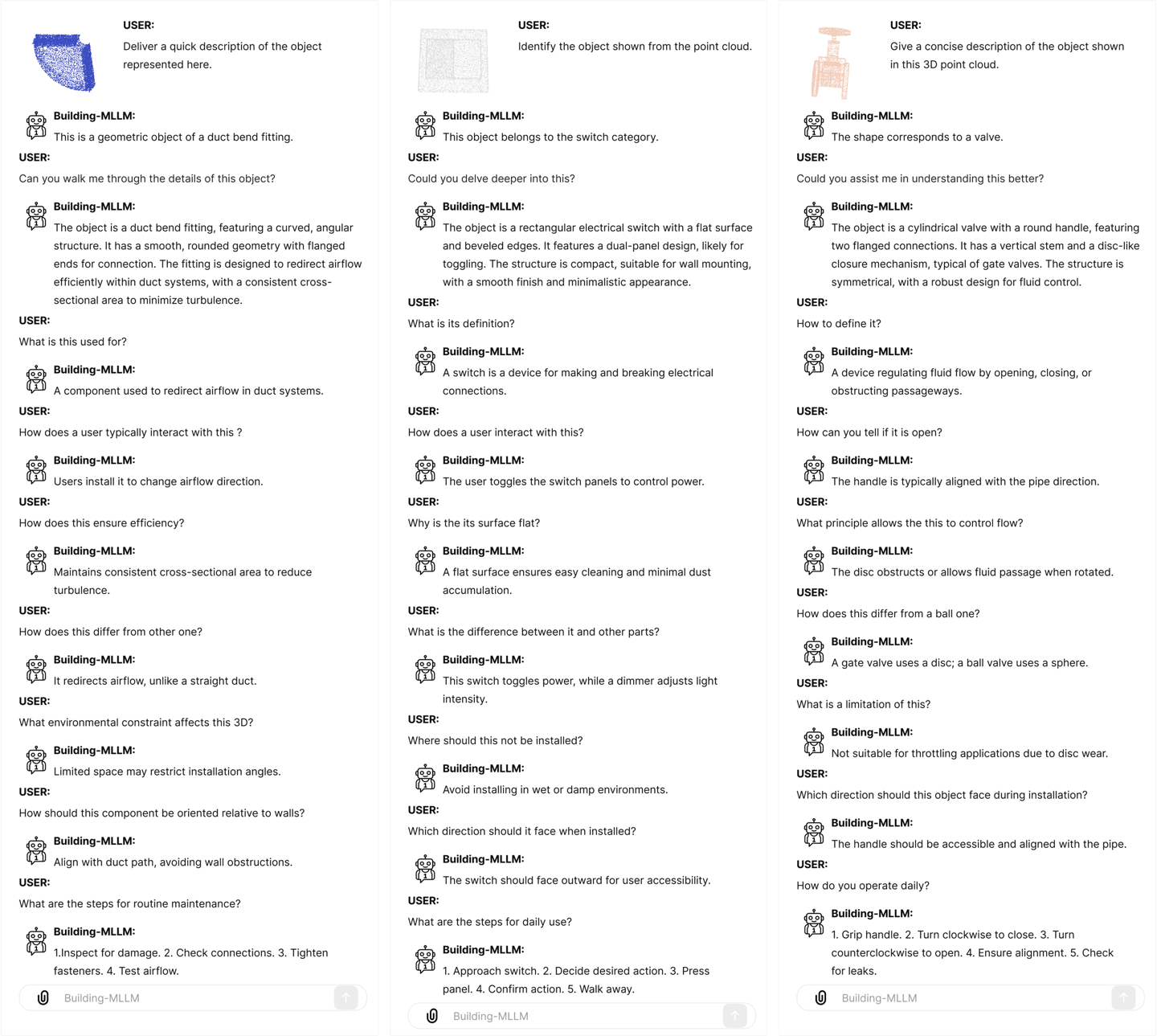}
\caption{Test case demonstration of Building-MLLM (2).}
\label{F9}
\end{figure}

\autoref{F8} and \autoref{F9} present qualitative test cases of Building-MLLM, demonstrating the model's comprehensive capabilities across multiple task dimensions. The model not only performs Simple Recognition and Complex captioning for diverse categories spanning the five major indoor systems, but also answers engineering knowledge questions that go beyond the caption content, and further engages in  multi-turn interactive instruction following, including constraint interpretation, spatial reasoning, structural logic inference, and embodied operation steps.

Together, these results form a coherent capability loop of geometry to language to engineering knowledge reasoning. On one hand, the model precisely binds three-dimensional geometric cues with linguistic expression, achieving consistent cross-modal understanding. On the other hand, it maintains semantic coherence, structural logic, and causal interpretability throughout multi-turn reasoning while exhibiting strong generalization and robustness in unseen scenarios. These characteristics indicate that Building-MLLM achieves unified competence ranging from fine-grained geometric perception to high-level engineering reasoning within complex indoor environments.

\newpage

\begin{figure}[!htb]
\centering
\includegraphics[width=\textwidth]{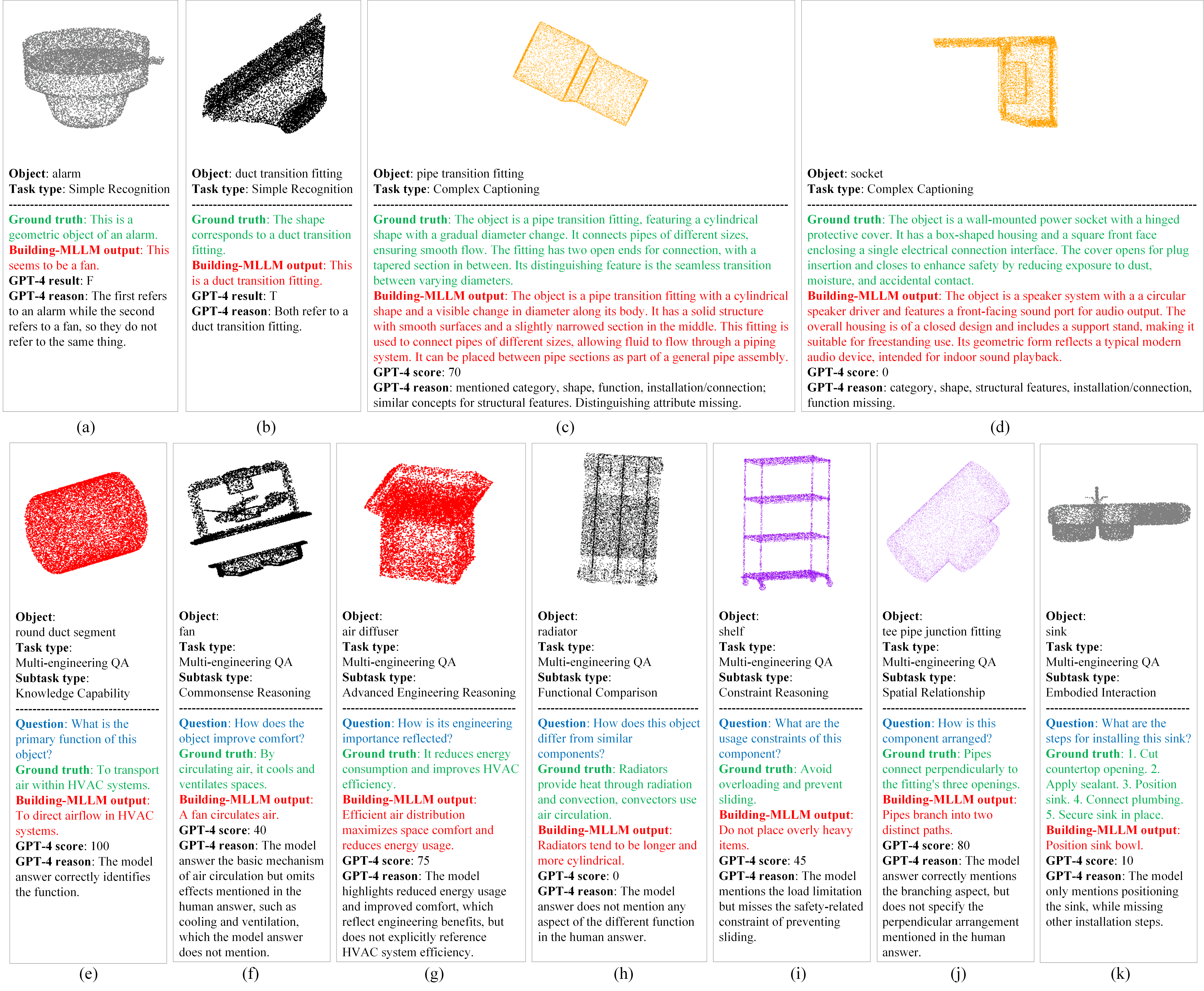}
\caption{Inference case studies with GPT-4–based evaluation.}
\label{F10}
\end{figure}

 \autoref{F10} presents more diverse examples covering Simple Recognition, Complex Captioning, and Multi-Engineering QA tasks, together with their GPT-4 evaluation results, demonstrating the transparency and reliability of the proposed evaluation process.As illustrated in (a) and (b), the Simple Recognition evaluation outputs binary T/F judgments along with explicit semantic category reasoning. For example, the evaluator determines correctness based on the semantic category "duct transition fitting" rather than the generic term "fitting". The (c) and (d) demonstrate evaluation results for the Complex Captioning task. Taking (c) as an example, the predicted description shows high consistency with the ground truth in terms of category, shape, function, and connection. However, it partially misses structural features such as open ends and lacks discriminative attributes like a seamless gradual transition. As a result, these missing semantic elements are systematically penalized under the structured semantic scoring scheme, leading to a final score of 70. The (e)–(k) present evaluation results for the Multi-Engineering QA task. In these cases, the scoring strictly depends on the specific question context and engineering constraints, effectively quantifying responses along a continuum ranging from completely incorrect to fully matched answers. This set of examples further demonstrates that the proposed evaluation framework provides interpretable and fine-grained assessment across diverse engineering-oriented tasks.

\newpage

To assess the potential transferability of the proposed Building-MLLM to real-world point clouds of indoor building components, we conduct several sets of inference experiments on three real-world point cloud datasets, aiming to preliminarily validate the model's capability for semantic understanding and instruction response generation on real component-level point clouds.

Specifically, we selected Pipework \citep{yeo2020deep}, ScanObjectNN \citep{uy2019revisiting}, and S3DIS \citep{armeni20163d} as three representative sources of real-world point clouds. Pipework is an MEP-oriented dataset that primarily consists of various pipe-related components, while ScanObjectNN is a general-domain real scanned object dataset that includes a subset of furniture and everyday objects. S3DIS, in contrast, is a typical real-world indoor scene dataset. Since large-scale real point cloud–text instruction-following datasets for indoor building components are still unavailable in the engineering domain, and constructing multi-task textual annotations for real point clouds is highly time-consuming and labor-intensive, this study adopts a limited-sample evaluation strategy to preliminarily validate the model's transferability to real-world data. Specifically, ten samples were selected from each of the three real-world point cloud datasets, resulting in a total of 30 real test samples. The sample selection was primarily empirical, aiming to cover different component categories, varying levels of geometric complexity, and different degrees of point cloud incompleteness, noise, or incomplete scanning. Subsequently, textual annotations were constructed for these samples across three tasks: Simple Recognition, Complex Captioning, and Multi-Engineering QA.

\autoref{TX} presents the transfer inference results of the proposed Building-MLLM and the original baseline PointLLM on multiple real-world point cloud datasets. Benefiting from the global token sampling mechanism in the point cloud encoder and the geometry-language alignment designs, including PIE, GPR, and the fixed textual prefix, Building-MLLM is still able to generate semantically coherent and engineering-oriented responses on real-world point cloud data. Compared with PointLLM, which does not incorporate the proposed architectural improvements and domain-adaptive training strategy, Building-MLLM achieves clearly superior GPT-based evaluation scores across all three tasks, including Simple Recognition, Complex Captioning, and Multi-engineering QA. Specifically, the overall scores are improved by 36.67\%, 31.49\%, and 38.46\%, respectively. These results indicate that the proposed synthetic-data-driven domain adaptation strategy endows Building-MLLM with a certain degree of generalisability and scalability when transferred to real-world indoor component point cloud datasets.

\begin{table}[!htb]
\footnotesize
\captionsetup{margin=0cm}
\centering
\caption{Comparison of transfer inference results between PointLLM and Building-MLLM on real-world point cloud datasets using GPT-4 evaluation.}
\label{TX}
\begin{tabular}{p{2.0cm} p{2.6cm} p{1.3cm} c c c}
\hline
Dataset & Method & Samples & 
Simple Recognition & Complex Captioning & Multi-engineering QA \\
\hline
\multirow{2}{*}{Pipework} 
& PointLLM & 10 & 20.00 & 15.18 & 9.60 \\
& Building-MLLM & 10 & 70.00 & 56.23 & 53.84 \\
\hline
\multirow{2}{*}{ScanObjectNN} 
& PointLLM & 10 & 40.00 & 24.47 & 14.80 \\
& Building-MLLM & 10 & 70.00 & 49.60 & 48.20 \\
\hline
\multirow{2}{*}{S3DIS} 
& PointLLM & 10 & 30.00 & 20.30 & 16.46 \\
& Building-MLLM & 10 & 60.00 & 48.58 & 54.20 \\
\hline
\multirow{2}{*}{Overall} 
& PointLLM & 30 & 30.00 & 19.98 & 13.62 \\
& Building-MLLM & 30 & 66.67 & 51.47 & 52.08 \\
\hline
\end{tabular}
\end{table}

\newpage

\begin{figure}[!htb]
\centering
\includegraphics[width=\textwidth]{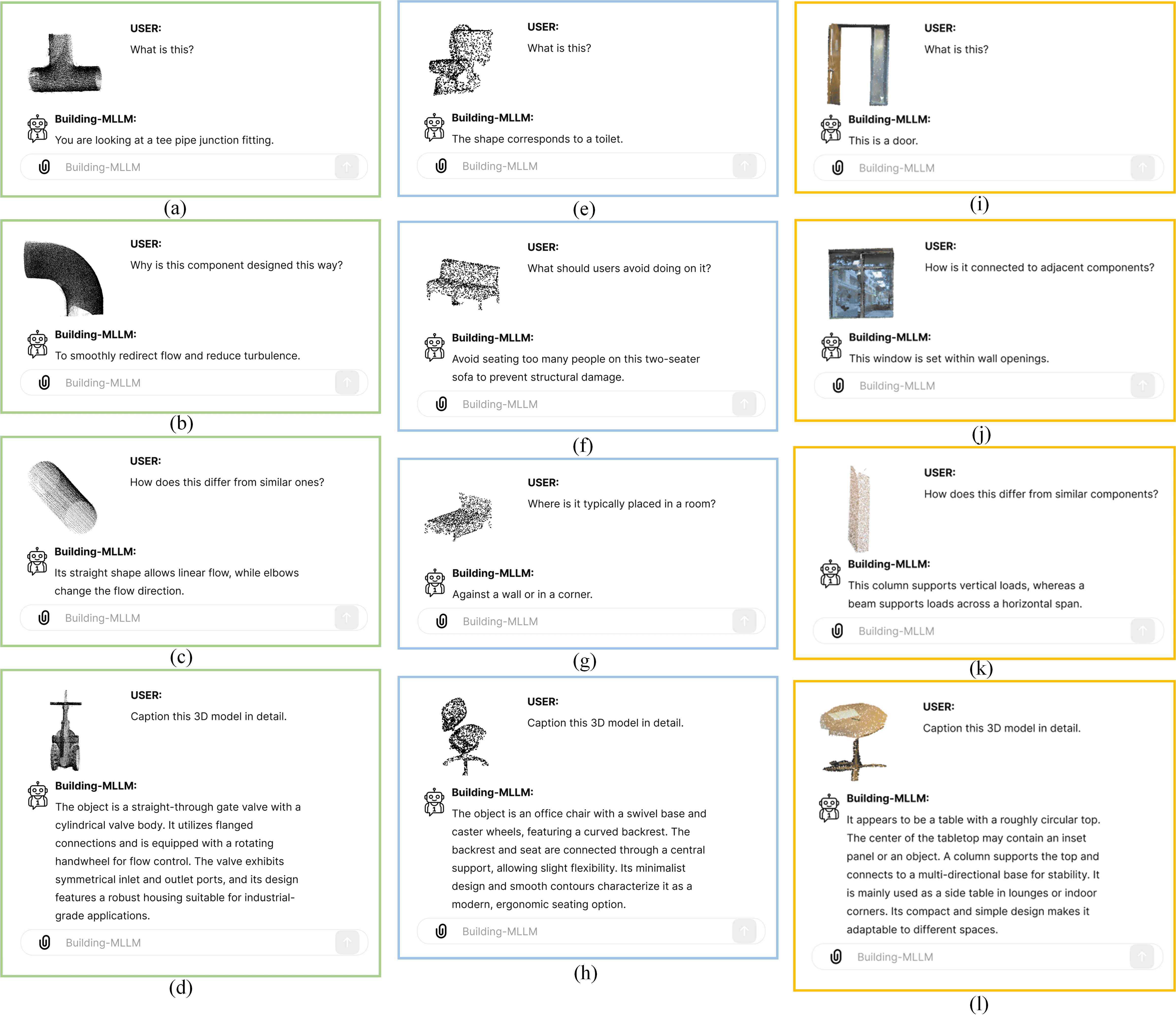}
\caption{Inference cases transferring Building-MLLM to real-world datasets.}
\label{F11}
\end{figure}

\autoref{F11}(a)--(d), (e)--(h), and (i)--(l) present representative inference cases of Building-MLLM on segmented point cloud clusters from Pipework, ScanObjectNN, and S3DIS, respectively, covering Simple Recognition, Complex Captioning, and Multi-Engineering QA tasks. Although real point clouds differ to some extent from synthetic ones in terms of point density distribution and geometric completeness, These results indicate that the proposed architectural and training strategies effectively enhance domain adaptation, and that such capabilities can be preliminarily verified on real-world point cloud datasets.

\newpage

\begin{figure}[!htb]
\centering
\includegraphics[width=\textwidth]{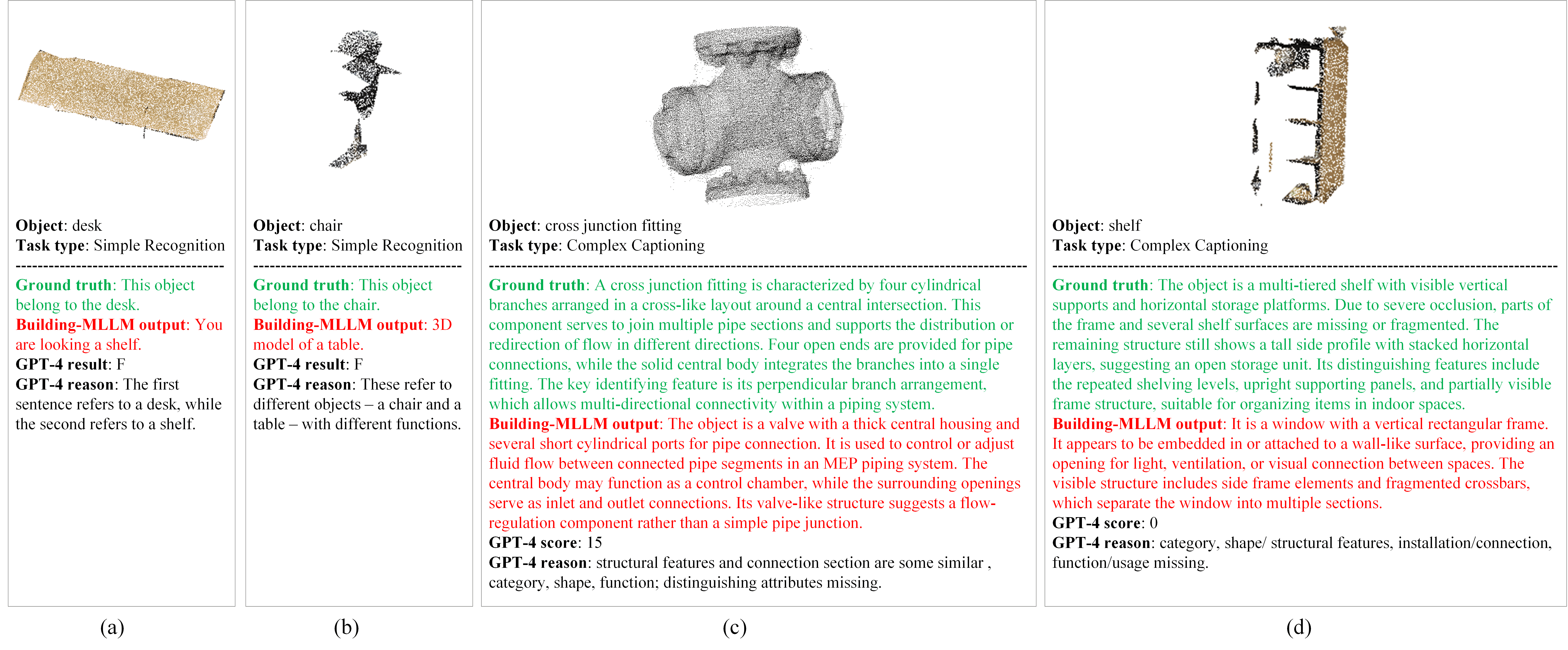}
\caption{Failure cases of Building-MLLM on real-world point cloud datasets.}
\label{F13}
\end{figure}

We also acknowledge that, although prior studies \citep{zhang2022pointclip,xue2023ulip} have shown the transferability of synthetic point clouds to real-world data, synthetic data cannot fully replace real point clouds. For example, compared with the inference results on the synthetic datasets (\autoref{T9}, \autoref{T10}, and \autoref{T11}), the inference results on the real-world datasets (\autoref{TX}) still show a certain degree of performance degradation across different metrics. The failure cases shown in the \autoref{F13} further demonstrate that, when real point cloud components suffer from severe incompleteness, non-typical shapes, or unclear category-specific features, the generalization ability of the proposed model remains limited. This also suggests that, although synthetic data can provide useful geometric priors and a semantic learning basis for the model, it still cannot fully cover the complexity and diversity of point cloud data in real engineering scenarios. Therefore, when sufficient domain-specific multimodal real-world data are available, models trained or jointly trained on large-scale real-world data are generally expected to achieve more robust and reliable performance. Future work will incorporate larger-scale real-world point cloud datasets and systematic point cloud--text annotations to support real-data-driven training and evaluation, thereby further validating the engineering applicability and generalization ability of the proposed method.

\subsection{Quantitative Analysis}

To systematically validate the effectiveness of Building-MLLM in both architectural design and parameter configuration, we conducted comprehensive quantitative experiments, including ablation studies and comparative evaluations.

In Stage 1, we focused on assessing the contribution of the geometry-language alignment, as its performance fundamentally determines the upper bound of subsequent instruction fine-tuning. Extensive ablation and comparison experiments were carried out in this stage and evaluated using the Simple Recognition task, demonstrating the model's superior domain adaptability and the advantages of the proposed alignment strategy. In Stage 2, we investigated a multi-dimensional LoRA optimization strategy. By evaluating the model across Simple Recognition, Complex Captioning, and Multi-Engineering QA, We systematically investigate how the proposed fine-tuning design, under the bidirectional geometry-language coupling mechanism, achieves balanced improvements across fine-grained recognition, long-range dependency modeling, and engineering-oriented reasoning.

\subsubsection{Stage-1 Alignment Performance}

 We employ GPT-4 under two instruction paradigms: the Instruction-typed prompt (``What is this?'') and the Completion-typed prompt (``This is an object of'') to enahnce evaluation reliability. As shown in \autoref{T12} (Experiments~1 and~2), Building-MLLM achieves alignment accuracies of 91.75\% and 91.25\% under the two prompt settings, respectively, consistently outperforming the baseline and yielding an average improvement of 7.00\%. To analyze the role of each component, we conduct single-factor ablation studies by individually removing modules from the full model. Using Experiment~2 as the reference, the results from Experiments~3--5 show that the fixed textual prefix, PIE, and GPR contribute average performance gains of 1.12\%, 5.25\%, and 2.12\%, respectively. Among them, PIE plays a central role by enhancing task-relevant discriminability and contextual modeling, while GPR and the fixed textual prefix provide complementary support by stabilizing geometric representations and guiding semantic alignment. Together, these components jointly address instruction robustness, domain-specific semantic enhancement, and geometric consistency, leading to a significant and robust performance improvement of Building-MLLM.

\begin{table}[!htb]
\footnotesize
\captionsetup{margin=0cm}
\caption{Comparative study of Building-MLLM after stage-1 training.}
\centering
\begin{tblr}{
  hline{1} = {-}{},
  hline{2,7} = {-}{black},
}
No. & Method & GPT(I) & GPT(C) & Avg. Score & $\Delta$ vs. 2 \\
1 & PointLLM & 84.75 & 84.25 & 84.50 & -7.00 \\
2 & \textbf{Building-MLLM} & \textbf{91.75} & \textbf{91.25} & \textbf{91.50} & -0.00 \\
3 & 2 w/o fixed textual prefix & 90.75 & 90.00 & 90.38 & -1.12 \\
4 & 2 w/o PIE & 86.50 & 86.00 & 86.25 & -5.25 \\
5 & 2 w/o GPR & 89.75 & 89.00 & 89.38 & -2.12 \\
\end{tblr}
\label{T12}
\end{table}

\begin{figure}[!htb]
\centering
\includegraphics[width=\textwidth]{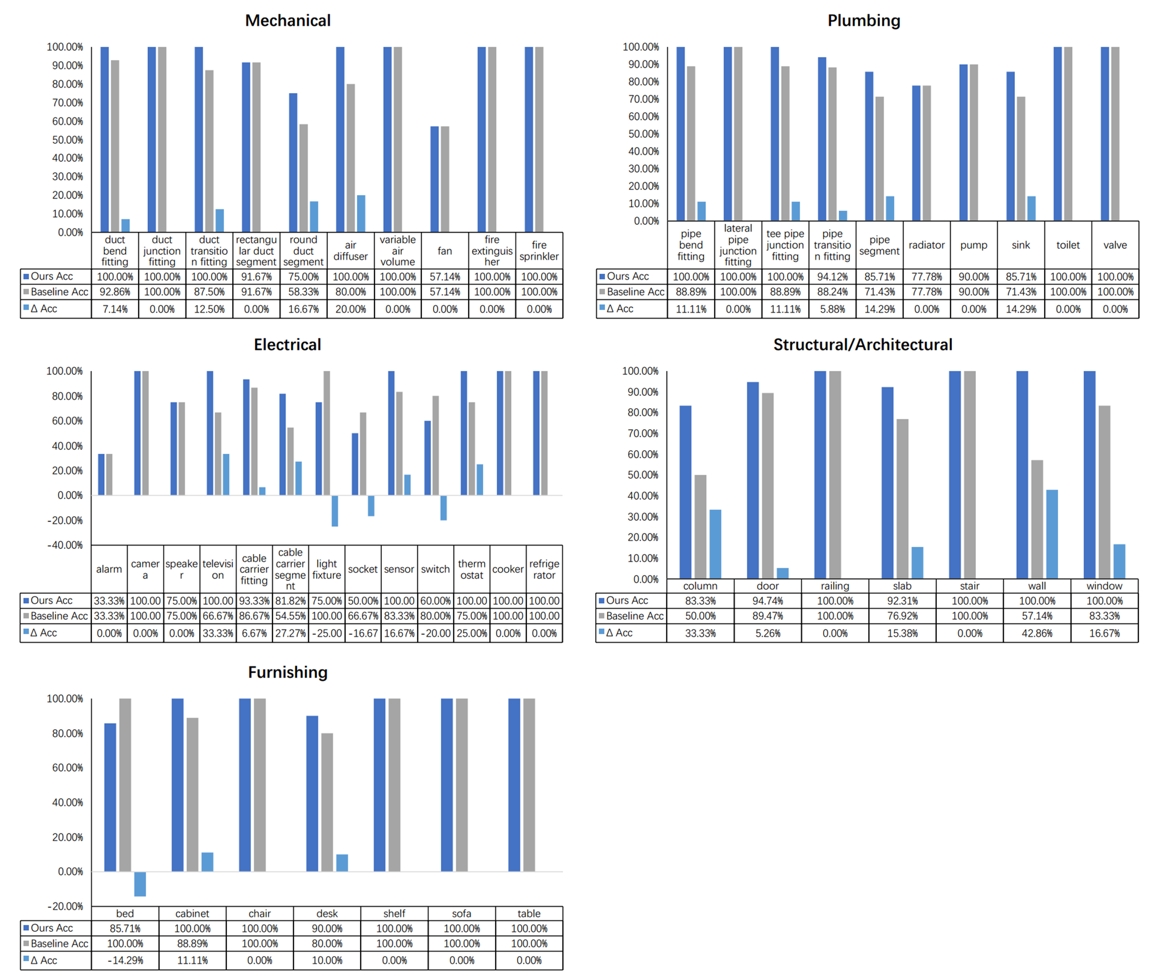}
\caption{Category-level recognition comparison.}
\label{F12}
\end{figure}

Furthermore, a category-level comparison between Experiment 1 and Experiment 2 by GPT(I) in \autoref{T12} was conducted. As illustrated in \autoref{F12}, among all 47 categories spanning five building systems, our method achieves performance improvements in 21 categories, maintains comparable performance in 22 categories, and exhibits degradation in only 4 categories when compared with the baseline. More specifically, in contrast to the Furnishing system, which is commonly involved in both professional engineering and general indoor environments, our method shows more pronounced and stable gains in systems with stronger engineering attributes, including Mechanical, Plumbing, Electrical, and Structural/Architectural categories. Consistent improvements are observed for components that require fine-grained local structural perception and strong geometric continuity, such as round duct segment and cable carrier segment, as well as for components that rely heavily on spatial consistency and discriminative geometric modeling, including wall and column. These results indicate that the proposed model is capable of effectively capturing subtle structural differences among components with similar geometric appearances, which is critical for engineering-oriented component understanding. By contrast, performance degradation is observed for a small number of categories (e.g., bed and light fixture). This can be attributed to the fact that such components appear frequently in large-scale generic indoor datasets and exhibit relatively consistent visual characteristics, enabling the baseline model to learn sufficiently robust representations from general-domain data. Under this condition, introducing feature enhancement strategies that emphasize engineering-oriented geometric modeling and structural awareness does not necessarily yield additional benefits for categories jointly dominated by generic and engineering semantics. Consequently, the observed degradation is reasonable and acceptable within the context of an overall engineering-focused understanding task.

\begin{table}[!htb]
\footnotesize
\captionsetup{margin=0cm}
\caption{Ablation study on the recursive learning of PIE.}
\centering
\begin{tblr}{
  hline{1} = {-}{},
  hline{2,7} = {-}{black},
}
No.  & Recursive Size & GPT(I) & GPT(C) \\
1 & 0              & 86.50   & 86.00   \\
2 & 1              & 90.50   & 89.00   \\
3 & 2              & 90.75   & 90.75   \\
4 & \textbf{3}     & 91.75   & 91.25   \\
5 & 4              & 91.25   & 91.00   
\end{tblr}
\label{T13}
\end{table}

\autoref{T13} reports the impact of the number of recursive learning within the PIE. The results show that, although the PIE substantially improves the alignment performance in Stage~1, increasing the number of recursive blocks does not lead to noticeable performance gains. This indicates that a  recursive block is already sufficient to provide effective feature enhancement. From the perspective of model complexity, one Recursive block contains approximately 742{,}080 parameters, accounting for only about 42\% of a single Point Encoder block (1{,}773{,}312 parameters). The entire encoder consists of 12 such blocks, with a total parameter scale on the order of \(2.1\times10^{7}\). Therefore, adding a small number of additional recursive blocks incurs only minimal overhead while modestly increasing the module's capacity, forming a sweet spot. We adopt three Recursive blocks as the configuration for the PIE.

As shown in \autoref{T14}, we further analyze the parameters design choices of the PIE. 
Experiments~1 and~2 indicate that removing semantic gating reduces the dynamic modulation capacity of positional encoding, leading to weaker alignment. For the frequency-encoding parameters \((\alpha,\beta)\), Experiment~3 employs \((1,1)\) as a no-decay, no-suppression baseline, while Experiment~4 adopts a higher-frequency variant \((3,1)\) to increase local detail sensitivity. Our final setting \((\alpha{=}2.5,\beta{=}0.8)\) provides a more favorable balance between frequency decay and amplitude control. Experiment~5 applies channel-wise modulation to the zero-initialized fusion parameter \(\gamma\). Although this finer granularity offers greater theoretical flexibility, the layer-wise alternative yields slightly better performance, and is therefore adopted as our default configuration.

\begin{table}[!htb]
\footnotesize
\captionsetup{margin=0cm}
\caption{Comparative analysis of key parameters in PIE.}
\centering
\begin{tblr}{
  hline{1} = {-}{},
  hline{2,7} = {-}{black},
}
No.  & PIE Design                  & GPT(I) & GPT(C) \\
1 & \textbf{Building-MLLM}        & 0.9175 & 0.9125 \\
2 & 1 w/o Gate                    & 0.8875 & 0.8775 \\
3 & 1 w $(\alpha{=}1,\ \beta{=}1)$      & 0.8775 & 0.8700 \\
4 & 1 w $(\alpha{=}3,\ \beta{=}1)$      & 0.8975 & 0.8975 \\
5 & 1 w channel-level $\gamma$    & 0.9125 & 0.9100 \\
\end{tblr}
\label{T14}
\end{table}

As shown in \autoref{T15}, unfreezing one to two shallow LLM layers under the GPR constraint yields the best alignment performance, while further unfreezing three or more layers leads to degradation.
This indicates that GPR benefits from limited shallow-layer unfreezing: moderate unfreezing injects essential geometric information without causing semantic drift in deeper layers, thereby improving geometry-language alignment while preserving linguistic stability.
This behavior is consistent with the hierarchical representation property of LLMs, where shallow layers are known to capture more local and structural features that are easier to align with geometric cues, whereas deeper layers primarily encode higher-level semantic abstractions that are more sensitive to disruption.

\begin{table}[!htb]
\footnotesize
\captionsetup{margin=0cm}
\caption{Sensitivity analysis of unfrozen shallow LLM layers under GPR.}
\centering
\begin{tblr}{
  hline{1} = {-}{},
  hline{2,7} = {-}{black},
}
No.  & Unfreeze LLM & GPT(I) & GPT(C) \\
1 & 0 layer        & 89.75   & 89.00   \\
2 & 1 layer        & 91.25   & 91.25   \\
3 & \textbf{2 layers} & 91.75   & 91.25   \\
4 & 3 layers       & 90.00   & 89.75   \\
5 & 4 layers       & 90.00   & 88.00   \\
\end{tblr}
\label{T15}
\end{table}

\subsubsection{Stage-2 Fine-Tuning Performance}

All multi-dimensional LoRA experiments are conducted based on a default setting, in which LoRA is applied to all 32 layers of LLaMA (Layer Range = All-32), injected into the Q/K/V/O projection modules, with a rank of 16 and a scaling factor of $\lambda = 32$.

In LoRA layer rank injection strategies, full-layer injection offers the complete coverage, last-16 layer injection focuses on high-level semantic generation, and even-index injection provides a structurally uniform pattern. \autoref{T16} shows that when these strategies are applied to domain-specific point cloud--text fusion and the unified modeling of tasks with varying sequence lengths, the performance of Building-MLLM becomes highly sensitive to the continuity and density of LoRA layer coverage, with the full 32-layer injection consistently achieving the best results.

\begin{table}[!htb]
\centering
\footnotesize
\captionsetup{margin=0cm}
\caption{Comparison experiments on LoRA layer ranges in LLaMA.}
\begin{tblr}{
  hline{1-2,5} = {-}{},
   colspec = {l| c| c| c c c c c c c c}
}
LLama & Rec & Cap & Know & CoSe & Engi & FuCp & Cstr & SpRe & EmIn &  Avg. \\
\textbf{All-32} & 91.00 & 61.27 & 80.35 & 62.97 & 72.4  & 67.03 & 59.56 & 58.62 & 37.18 & 62.58 \\
Last-16 & 87.75 & 48.29 & 58.08 & 33.23 & 37.44  & 40.19 & 21.69 & 37.46 & 19.49 & 35.36 \\
Even-Idx &  88.89 & 58.90& 75.86  & 57.61  & 64.47  & 62.11  & 55.32  &  52.88  &  30.80  &  57.00    &        
\end{tblr}

\begin{flushleft}
\footnotesize\textit{Abbreviations:}
\textit{Rec} = Simple Recognition;\ 
\textit{Cap} = Complex Captioning;\ 
\textit{Know} = Knowledge Capability;\ 
\textit{CoSe} = Commonsense Reasoning;\ 
\textit{Engi} = Advanced Engineering Reasoning;\ 
\textit{FuCp} = Functional Comparison;\ 
\textit{Cstr} = Constraint Reasoning;\ 
\textit{SpRe} = Spatial Relationship;\ 
\textit{EmIn} = Embodied Interaction.
\end{flushleft}

\label{T16}
\end{table}

This outcome can be explained by how different injection schemes reshape the geometry--language fusion pathway inside the mixed point-cloud--text sequence. Injecting LoRA only into the final 16 layers primarily affects high-level linguistic decoding and cross-sentence coherence, yet it cannot account for the shallow- and mid-layer interactions where geometric cues and textual semantics are first integrated. Even-index injection achieves moderate improvement, but its intermittent structure disrupts the sequential continuity required for stable fusion. Although stage~1 provides an initial alignment between geometric and textual representations, the multi-task nature of Building-MLLM further demands support for both short sequences that rely on rapid geometric grounding and long sequences that require multi-step reasoning. Ensuring reliable performance across this spectrum of task demands necessitates preserving smooth, layer-spanning geometry--language coupling, which explains why full-layer injection consistently remains the most effective configuration.

\begin{table}[!htb]
\centering
\footnotesize
\captionsetup{margin=0cm}
\caption{Comparison experiments on LoRA rank r in LLaMA.}
\begin{tblr}{
  hline{1-2,5} = {-}{},
  colspec = {l| c |c| c c c c c c c c}
}
r & Rec & Cap & Know & CoSe & Engi & FuCp & Cstr & SpRe & EmIn &  Avg. \\
8  & 90.50 & 60.62  & 78.8  & 60.5  & 68.78 & 64.44 & 56.09 & 56.66 & 32.58 & 59.69 \\
\textbf{16} & 91.00 & 61.27 & 80.35 & 62.97 & 72.4  & 67.03 & 59.56 & 58.62 & 37.18 & 62.58 \\
32 & 87.25 & 63.21 & 81.48 & 67.28 & 73.6  & 67.91 & 61.19 & 63.03 & 45.49 & 65.71 \\
\end{tblr}
\label{T17}
\end{table}

As shown in \autoref{T17}, increasing the LoRA rank \(r\) from 8 to 16 and 32 leads to a rise-then-fall trend in Building-MLLM's performance on the Simple Recognition task, whereas performance continues to improve on Complex Captioning and Multi-Engineering QA task. This pattern indicates that higher ranks provide stronger adaptation capacity, enabling the model to better capture cross-sentence dependencies, factual integration, and complex reasoning chains. However, they also amplify the influence of high-level parameter updates on shallow lexical structure and geometric grounding. Overall, \(r=16\) is better suited for short-range tasks involving recognition while \(r=32\) offers advantages for tasks that rely heavily on long-range semantic dependencies and reasoning.

\begin{table}[!htb]
\centering
\footnotesize
\captionsetup{margin=0cm}
\caption{Comparison experiments on LoRA target modules in LLaMA.}
\begin{tblr}{
  hline{1-2,6} = {-}{},
    colspec = {l| c| c| c c c c c c c c}
}
Target modules & Rec & Cap & Know & CoSe & Engi & FuCp & Cstr & SpRe & EmIn & Avg. \\
q/v         & 89.00 & 60.00 & 78.97 & 62.43 & 69.30 & 64.78 & 56.48 & 56.72 & 34.43 & 60.44 \\
q/v/k/o     & 91.00 & 61.27 & 80.35 & 62.97 & 72.40 & 67.03 & 59.56 & 58.62 & 37.18 & 62.58 \\
\textbf{q/v/k/o/u/d} & 88.25 & 65.26 & 83.42 & 68.94 & 74.31 & 71.26 & 65.86 & 66.97 & 50.89 & 68.80 \\
q/v/k/o/u/d (r=32)     & 84.25 & 63.05 & 82.70 & 66.96 & 74.83 & 71.06 & 64.82 & 67.39 & 52.44 & 68.59 \\
\end{tblr}
\label{T18}
\end{table}

As shown in \autoref{T18}, expanding LoRA injection from \(\{q,v\}\) to \(\{q,v,k,o\}\) consistently improves performance across tasks, establishing a more stable geometry--language fusion pathway and enhances semantic and geometric consistency across tokens. Further injecting LoRA into the FFN sub-layers \(\{u,d\}\) yields larger gains in Complex Captioning and Multi-Engineering QA, as FFN layers strengthen deep semantic transformations and cross-sentence information integration. However, this also weakens shallow lexical and geometric binding, resulting in degraded performance on Simple Recognition, which relies on short-range cues. When the rank increases to \(r=32\), all tasks show performance drops, indicating that this injection–rank combination reaches a capacity boundary where further expansion leads to overfitting and semantic drift. We therefore adopt the \(\{q,v,k,o,u,d\}\) configuration with \(r=16\) as the setting for subsequent experiments.

\begin{table}[!htb]
\centering
\footnotesize
\captionsetup{margin=0cm}
\caption{Comparison experiments on LoRA scaling strategies in LLaMA.}
\begin{tblr}{
  hline{1-2,5} = {-}{},
   colspec = {l| c| c| c c c c c c c c}
}
scaling strategies & Rec & Cap & Know & CoSe & Engi & FuCp & Cstr & SpRe & EmIn &  Avg. \\
Uniform Scaling($\lambda = 32$) & 88.25 & 65.26 & 83.42 & 68.94 & 74.31  & 71.26 & 65.86 & 66.97 & 50.89 & 68.80 \\
\textbf{Layer-wise Progressive Scaling} & 88.75 & 64.92 & 80.93 & 71.35 & 73.68 &  68.86 & 66.87 &  69.65 & 53.30&   69.27    \\
Module-wise Differential Scaling & 87.00  & 65.34  & 85.84  & 69.11  & 77.45  &  72.35  &  62.24  &  65.42  &  45.08  &  68.21      
\end{tblr}
\label{T19}
\end{table}

As shown in \autoref{T19}, we further explore balance by adjusting the LoRA scaling coefficient \(\lambda\). Two variants are examined: a layer-wise progressive scaling scheme (16 for the first 8 layers, 32 for the middle 16 layers, and 64 for the last 8 layers) and a module-wise differential scaling scheme (24 for \(q,v\) and 48 for \(o,k,u,d\)). The results show that layer-wise progressive scaling delivers stable gains on Simple Recognition and Multi-Engineering QA, with the latter benefiting across all four sub-tasks. Although Complex Captioning slightly declines, this indicates that a conservative shallow, enhanced deep distribution better preserves model stability while strengthening high-level semantic integration and cross-sentence reasoning, achieving a more balanced multi-task trade-off. In comparison, the module-wise differential scheme provides only marginal improvements in Complex Captioning and consistently degrades Simple Recognition and Multi-Engineering QA, demonstrating weaker overall consistency. Thus, layer-wise progressive scaling offers a superior balance and surpasses the uniform-scaling baseline.

\section{Conclusion}
\label{conclusion}
This paper addresses the long-standing absence of systematic solutions for linguistic understanding of indoor building-component point clouds by proposing Building-MLLM, a MLLM framework that deeply integrates point cloud geometry with language interaction. The paper presented the development of a progressive instruction-following generation engine, capable of automatically producing three types of engineering-constrained instruction-following data (Simple Recognition, Complex Captioning, and Multi-Engineering QA). Based on this engine, we compile the first indoor building component point cloud-text instruction-following dataset, covering five systems, 47 categories, 4,198 objects, and 37,782 instruction-following pairs. To achieve high-quality geometry--language alignment, we introduce a three-part alignment mechanism consisting of a fixed textual prefix, PIE, and GPR: the fixed prefix stabilizes linguistic patterns and suppresses semantic drift, PIE enhances task-relevant geometric representations, and GPR constrains shallow geometric features to mitigate structural erosion during high-level semantic updates. In addition, we design a multi-dimensional LoRA optimization strategy tailored for mixed point cloud--text sequences and multi-task learning, enabling controllable performance balancing across layers, modules, ranks, and scaling factors. Finally, beyond automatic metrics, we introduce GPT-4 as a semantic evaluator equipped with task-specific evaluation prompts. Together, these components form a complete pipeline---from data generation and geometric--linguistic alignment to instruction tuning and evaluation---laying the methodological foundation for point-cloud-driven multimodal intelligence in building operation and maintenance scenarios.

Experimental results demonstrate substantial improvements across all stages of the proposed framework. 
In the standalone alignment evaluation, the first-stage Building-MLLM achieves accuracies of 
91.75\% (instruction-typed) and 91.25\% (completion-typed), outperforming PointLLM by approximately 
7.00\%, confirming the effectiveness of the fixed prefix, PIE, and GPR in strengthening geometry--language 
consistency. In second-stage instruction tuning, the proposed multi-dimensional LoRA strategy 
successfully balances short-range tasks (Simple Recognition) and long-range tasks (Complex Captioning 
and Multi-Engineering QA), achieving more stable and robust performance in mixed-sequence multi-task 
settings. On the independent test-inference set evaluated with GPT-4, Building-MLLM obtains scores of 
88.00\%, 65.10\%, and 68.14\% across the three task types, significantly surpassing PointLLM's 81.50\%, 
58.40\%, and 59.57\%, thereby validating the model's full-chain cross-modal capability from geometric 
understanding to semantic generation and multi-sentence reasoning.
Moreover, under a limited-sample evaluation on three real-world datasets, Building-MLLM improves the transfer inference scores over the original PointLLM baseline by 36.67\%, 31.49\%, and 38.46\% across the three tasks, respectively, indicating a certain degree of generalisability and scalability in real-world indoor component understanding.

Several limitations warrant attention. The synthetic dataset provided the annotation quality needed to establish the methodology, but operational facilities introduce complexity such as occlusion, corrosion, and non-standard components that current training does not capture. Although we conducted inference on limited real-world datasets to demonstrate transfer feasibility, achieving large-scale and reliable coverage will require systematic domain adaptation and the support of comprehensive multimodal datasets. In addition, although the proposed dataset was validated through manual reliability assessment, the reviewers were mainly PhD students with similar academic backgrounds, which may introduce a potential risk of shared bias. Finally, our study focuses on component-level understanding and does not yet support system-level reasoning over connectivity, spatial context, and global constraints, which are essential for holistic facility management and digital twin applications.

Future work will address these limitations. First, incorporating large-scale real facility scans with domain adaptation techniques will help bridge the synthetic-to-real gap while building on the methodological foundation established here. Reviewer diversity will also be expanded to include broader domain expertise and industry experience, thereby improving the reliability and generalizability of dataset evaluation. Second, integrating multi-sensor fusion (RGB-D, thermal imaging) and retrieval-augmented generation with external regulatory knowledge sources will enable finer-grained, physically grounded semantics and provide domain-compliant responses for facility management workflows. Third, developing scene-level understanding modules will enable system connectivity reasoning and spatial context aggregation, transforming our Building-MLLM from a component-level understanding unit into a comprehensive facility intelligence platform that supports proactive O\&M decision-making.

\appendix

\section*{Declaration of Generative AI in Scientific Writing}
During the preparation of this work, the authors used OpenAI's tool, ChatGPT in order to proofread the manuscript. After using this tool, the authors reviewed and edited the content as needed and took full responsibility for the content of the publication.

\section*{Acknowledgments}
This work was supported by the China Postdoctoral Science Foundation (No. 2023M740761), the National-Local Joint Engineering Laboratory on Digital Preservation and Innovative Technologies for the Culture of Traditional Villages and Towns(2025HSKFJJ004, 2025HSKFJJ009), the GDAS' Special Project of Science and Technology Development (2023GDASQNRC-0216, 2022GDASZH-2022010111),  and the Science and Technology Program of Guangdong(2024B1212080002)

\section*{Declaration of Competing Interest}
The authors declare that they have no known competing financial interests or personal relationships that could have appeared to influence the work reported in this paper.

\section*{Data Availability}

The code is publicly available at: 
\url{https:github.com/JingShuju/Building-MLLM}.



\bibliographystyle{elsarticle-num}
\bibliography{references} 

\section{Prompt Robustness Analysis}

During the training stage, we introduced a diverse set of natural-language instructions to enhance the model's adaptability to different textual expressions. To further verify the robustness of the model to varying natural-language instructions at the testing stage, namely that its performance primarily depends on task semantics rather than the literal wording of specific instructions, we conducted a prompt-robustness consistency evaluation as a sanity check.

Specifically, under the strict condition that the test set, model parameters, and inference configurations were kept completely identical, we evaluated the model using three additional instructions that are semantically equivalent but linguistically different, randomly selected from the instruction set used during training, in addition to the default unified prompt (e.g., "What is it?"). The experimental results are summarized in the \autoref{TP1}:

\begin{table}[!htb]
\centering
\footnotesize
\captionsetup{margin=0cm}
\caption{Prompt-robustness consistency evaluation.}
\begin{tblr}{
  colspec={>{\color{black}}l >{\color{black}}c},
  hline{1} = {-}{},
  hline{2,6} = {-}{black},
}
Instruction                                         & Accuracy (\%) \\
What
  is it?                                       & 88.00         \\
Describe
  the 3D point cloud object briefly.        & 87.50         \\
Give
  a concise description of this 3D point cloud & 87.50         \\
Identify.
  the object shown from the point cloud.   & 88.25         
\end{tblr}
\label{TP1}
\end{table}

It can be observed that under the four different prompt settings, the recognition accuracy remains stable within the range of 87.50\% to 88.25\%, with only minor variation. This indicates that the model's recognition performance does not rely on the literal form of any specific prompt, but instead responds consistently to the core task semantics of object identification from point clouds.

These results further validate the effectiveness of the multi-instruction learning strategy introduced during training. This strategy enhances the model's adaptability to variations in natural-language expressions and improves prediction stability at the inference stage under different but semantically equivalent prompts. Moreover, the use of multiple prompts for evaluation better reflects realistic application scenarios, in which users may issue recognition requests using diverse natural-language expressions.

\end{document}